\pdfoutput=1
\documentclass[11pt]{article}
\usepackage[utf8]{inputenc}
\usepackage{newunicodechar}
\newunicodechar{₂}{$_2$}
\newunicodechar{∗}{\ensuremath{*}}
\usepackage[final]{acl}
\usepackage{array}
\usepackage{times}
\usepackage{latexsym}
 \usepackage{amsmath}
\usepackage{amssymb}
\usepackage[T1]{fontenc}
\usepackage{booktabs}        % \toprule, \midrule, \bottomrule 등 표의 품질 향상
\usepackage{multirow}        % \multirow 명령어 사용 가능하게 함
\usepackage{graphicx}        % \resizebox 사용을 위해 필요
\usepackage{array}           % \arraybackslash 등을 사용하기 위해 필요
\usepackage{caption}         % 표/그림 캡션 설정 관련 (선택적이지만 권장)
\usepackage{xcolor}          % \textcolor 사용을 위해 필요 (회색 ± 값)

\usepackage[utf8]{inputenc}

\usepackage{microtype}
\usepackage{multirow}
\usepackage{colortbl} 
\usepackage{inconsolata}

\usepackage{graphicx}

\usepackage{comment}
\usepackage{framed}
\usepackage{tcolorbox}
\tcbuselibrary{skins,breakable}
\usepackage{xcolor}
\usepackage{pifont}
\newcommand{\cmark}{\ding{51}}
\newcommand{\xmark}{\ding{55}}
\usepackage{booktabs}
\usepackage[dvipsnames]{xcolor}
\usepackage{tablefootnote}
\usepackage{threeparttable}

\usepackage{tabularx}
\usepackage{booktabs}
\usepackage{multirow}

\usepackage{calc}
\newtcolorbox{promptbox}[1][]{
  colback=gray!5!white,      % 연한 회색 배경
  colframe=gray!50!black,    % 진한 회색 테두리
  boxrule=0.5pt,             % 테두리 두께
  arc=2mm,                   % 모서리 둥글게
  fontupper=\ttfamily\small, % 내용 폰트 (타자체, 작게)
  title=\textbf{ }, % 박스 제목
  coltitle=white,
  #1
}

\definecolor{paperbg}{HTML}{F6F8FA}
\definecolor{paperline}{HTML}{1F6FEB}
\definecolor{good}{HTML}{1A7F37}

\newtcolorbox{paperbox}[1][]{
  enhanced,
  colback=paperbg,
  colframe=paperline,
  boxrule=0pt,
  borderline west={2pt}{0pt}{paperline},
  arc=1.5mm,
  left=3mm,right=3mm,top=2mm,bottom=2mm,
  fontupper=\small,
  breakable,
  #1
}
\title{\textsc{REPAIR}: Resolving Long-Tail Confusion in Scientific Retrievers\\ via Fact-Verified Iterative Refinement}

\author{
  \textbf{Yerim Oh\textsuperscript{1}} \quad
  \textbf{Gunhee Kim\textsuperscript{1}}
\\
  \textsuperscript{1}Seoul National University
\\
  \texttt{yerim.oh@vision.snu.ac.kr, gunhee@snu.ac.kr}}
\definecolor{myhighlight}{HTML}{ EEDCDC}
\newcommand{\myhl}[1]{{\setlength{\fboxsep}{0.5pt}\colorbox{myhighlight}{#1}}}

\begin{document}
\maketitle
\begin{abstract}
Precise retrieval of scientific information is fundamentally constrained by \textit{long-tailed concepts} and \textit{high fact-sensitivity} of scientific corpora. These challenges often limit the effectiveness of dense retrievers and hallucination-prone LLM augmentation. To address this, we present \textsc{REPAIR}, a self-evolving data augmentation framework for scientific dense retrievers. \textsc{REPAIR} iteratively synthesizes training data to address knowledge gaps by cycling through 
diagnosis of long-tail concepts, API-guided evidence expansion, and differentiation via hard negative mining. This process effectively grounds retrieval in factual reality to resolve fine-grained distinctions. Extensive experiments demonstrate that \textsc{REPAIR} significantly outperforms 19 strong baselines on nine materials science and biomedical benchmarks. Our work highlights that diagnosing and factually augmenting data to long-tail deficits is essential for robust scientific retrieval.\footnote{Our code is available at \url{https://github.com/yerimoh/REPAIR}}
\end{abstract}

\section{Introduction}

In highly specialized fields such as materials science and biomedicine, the continuous influx of new literature makes efficient knowledge discovery a critical challenge \cite{sharma2025og, choudhary2022recent, kononova2021opportunities}. To address this, retrieval-augmented generation (RAG) has emerged as a promising methodology to dynamically incorporate up-to-date domain knowledge. By grounding generation in precise information retrieval (IR), RAG enables reliable downstream applications, including question answering \cite{sohn2025rationale, zhang2024honeycomb}, knowledge discovery \cite{ocana2025integrating, pei2025language}, and scientific decision-making \cite{chiang2025llamp, ong2025large}. The success of these applications fundamentally depends on the accuracy of the underlying IR models.

However, while state-of-the-art dense retrievers excel on general-domain text, they suffer significant performance degradation when applied to scientific corpora. This lexical and semantic gap is widely recognized as the domain shift issue \cite{kamalloo:2024}. To mitigate this, recent studies have adapted retrievers by aggregating domain-specific datasets or utilizing synthetic data augmentation \cite{jin2023medcpt, zhang2023pre, singh2023scirepeval, xu2024bmretriever}. Despite yielding empirical improvements, these approaches largely treat scientific documents as standard text, overlooking the intrinsic characteristics that distinguish scientific literature from general domains.

\begin{figure}[!t] % Table에서 Figure로 변경
\centering
\scriptsize

\definecolor{ltpink}{HTML}{FADBD8} % 톤다운된 연한 
\definecolor{ltgreen}{HTML}{CBE8D7} % 톤다운된 연한 초록 (REPAIR 팩트)
\definecolor{ltMONE}{HTML}{   EBF2F6} % 톤다운된 연한 초록 (REPAIR 팩트)

% 행간이 벌어지는 것을 막기 위해 하이라이트 박스 여백(fboxsep) 최소화
\newcommand{\hlMINE}[1]{{\setlength{\fboxsep}{0.1pt}\colorbox{ltMONE}{#1}}}
\newcommand{\hlbad}[1]{{\setlength{\fboxsep}{0.1pt}\colorbox{ltpink}{#1}}}
\newcommand{\hlgood}[1]{{\setlength{\fboxsep}{0.1pt}\colorbox{ltgreen}{#1}}}
\setlength{\tabcolsep}{1.5pt}
\renewcommand{\arraystretch}{1.3} % 필요시 1.1이나 1.0으로 줄이셔도 됩니다.

\begin{tabular}{>{\centering\arraybackslash}m{0.13\linewidth} | m{0.8\linewidth}}
\hline
\textbf{Method} & \textbf{Augmented Pairs} \\ \hline
\textbf{LLM} \par \textbf{Gen.} \par (\citeyear{xu2024bmretriever}) & 
\textbf{Query:} Which myeloma cell line carries prosurvival \hlbad{\textit{BCL1}}? \vspace{0.06cm} \newline
\textbf{Positive Document:} \ldots Myeloma cells usually express a range of the prosurvival \textit{BCL2} proteins \ldots \vspace{0.15em}
\\

\hline
\textbf{REPAIR} (ours) & 
\textbf{Query:} The role of \hlgood{\textit{BCL2}} and its pro-survival relatives in tumourigenesis and cancer therapy \vspace{0.06cm} \newline
\textbf{Positive Document:} \ldots In this invited review article, we reminisce on the discovery of \hlgood{\textit{BCL2}}, we discuss mechanisms \ldots
\vspace{0.06cm}  \\
\hline
\end{tabular}

\vspace{0.5em}
\centerline{\small (a) Comparison of augmented pairs}
\vspace{0.1em}

% --- (b) 그래프 부분 (아래로 이동) ---
\begin{minipage}{0.48\linewidth}
    \centering
    \includegraphics[width=\linewidth]{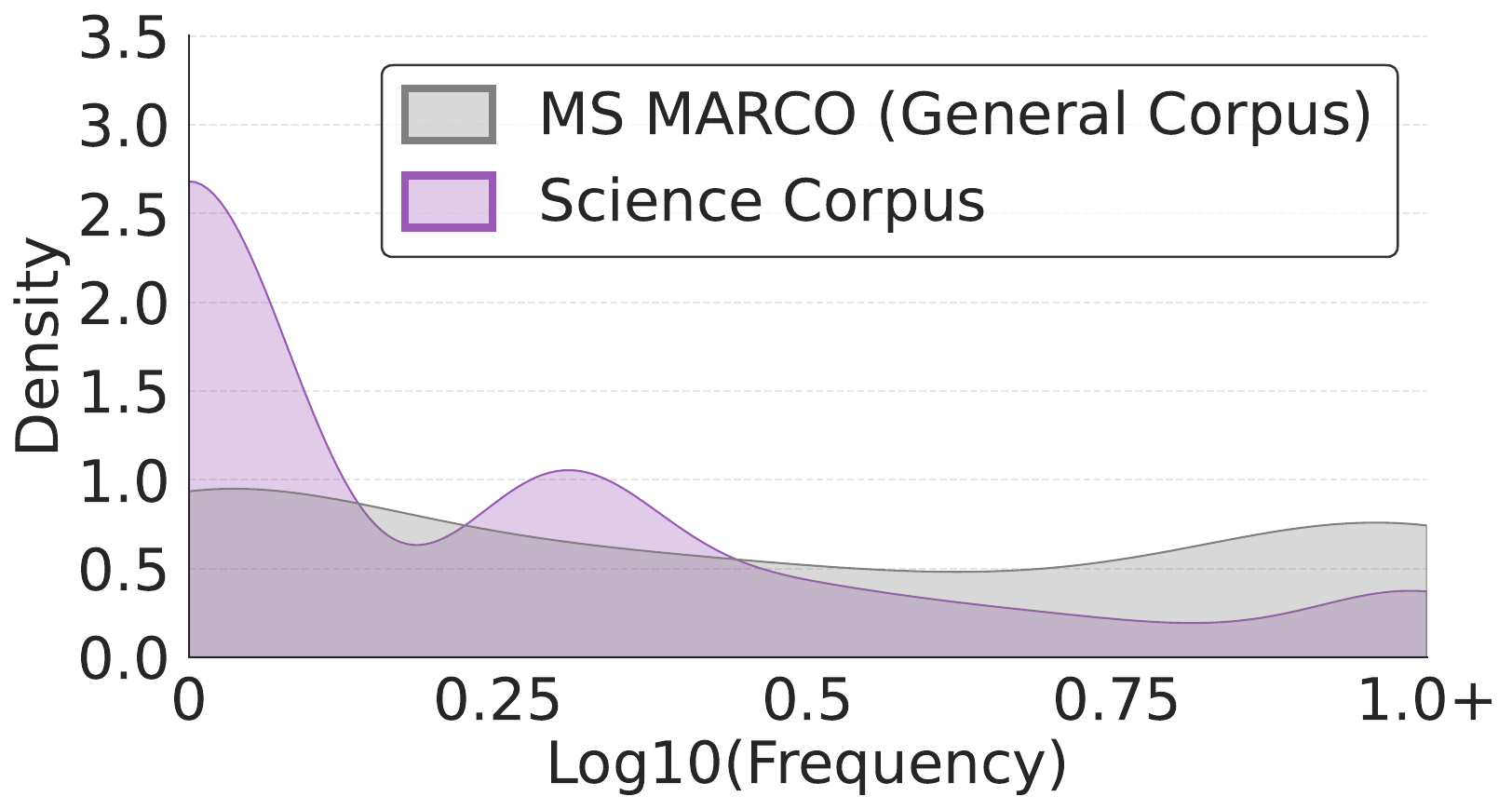}\vspace{-0.1em} \\
    \scriptsize Domain-level Long-tail
\end{minipage}
\hfill
\begin{minipage}{0.48\linewidth}
    \centering
    \includegraphics[width=\linewidth]{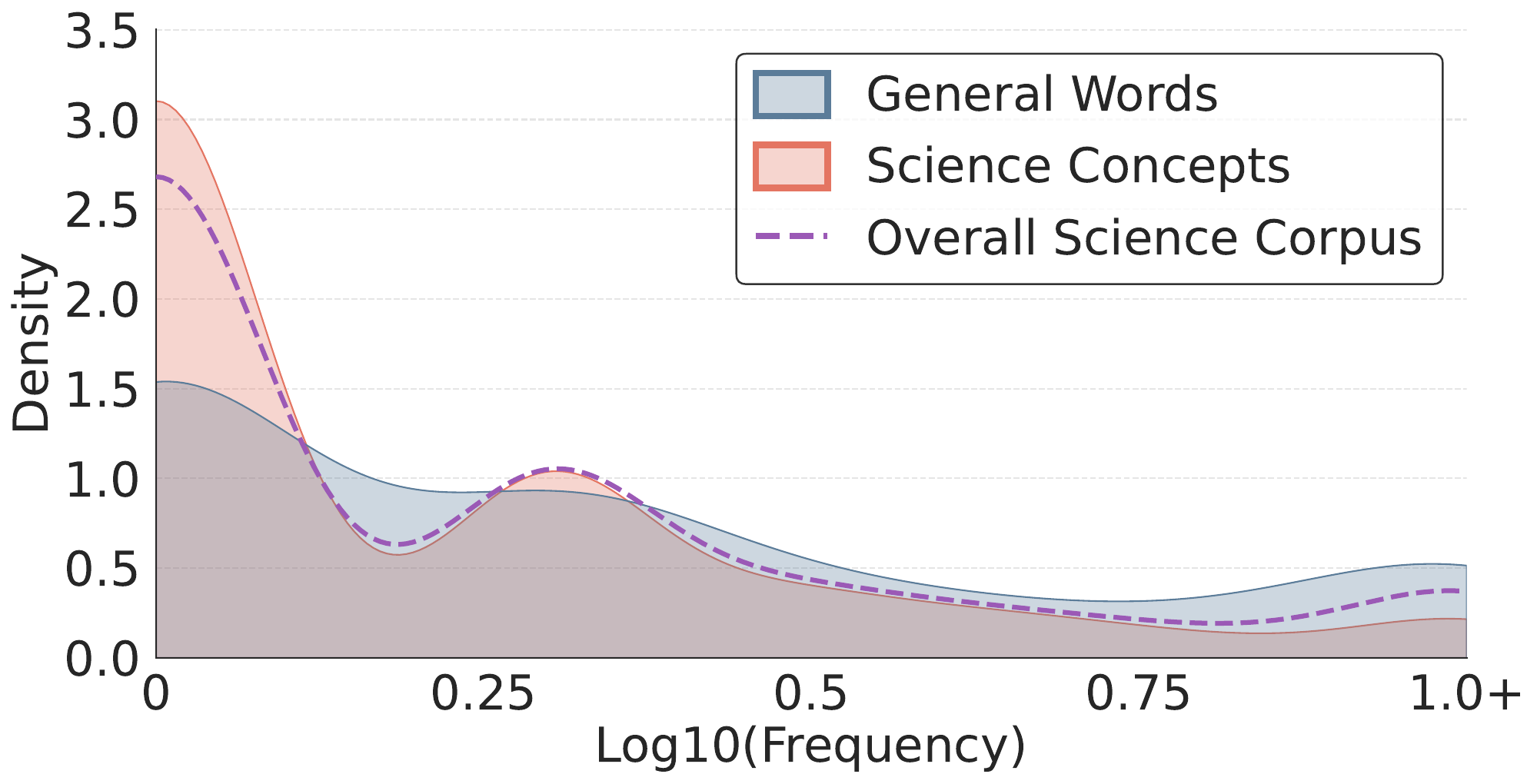}\vspace{-0.1em} \\
    \scriptsize Concept-level Long-tail
\end{minipage}

\vspace{0.5em}
\centerline{\small (b) Long-tail distributions}

% 캡션 보강 (리뷰어 피드백 반영)

\caption{Motivation of REPAIR. (a) While existing data augmentation methods generate structural \hlbad{hallucinations} (e.g.,  lexically similar \textit{BCL1} and  \textit{BCL2} critically corrupt scientific facts), REPAIR accurately grounds \hlgood{condition-sensitive scientific facts}. (b) Log-frequency analysis on scientific vs. general corpora reveals that the severe long-tail in scientific domains (left) is predominantly driven by \textit{scientific concepts} (right).
}
\label{fig:intro_overview}
\end{figure}

Specifically, scientific retrieval is governed by two structural properties: \textbf{long-tailed concept distribution} and \textbf{high fact-sensitivity}. Scientific corpora exhibit extreme long-tail distributions composed of irreplaceable entities such as chemical formulas, rare molecular structures, and specific gene or protein families \cite{oh-etal-2025-incorporating}. Unlike general-domain terms, these entities lack semantic substitutes, making naive data augmentation ineffective. As shown in Figure~\ref{fig:intro_overview}, specific proteins like \textit{BCL2} represent such entities that cannot be loosely generalized. A multi-corpus statistical characterization of this long tail is given in Appendix~\ref{app:longtail_stats}. Moreover, scientific outcomes are hypersensitive to precise terminology and experimental conditions. A single-character hallucination from \textit{BCL2} to \textit{BCL1} invalidates the generated query. Such plausible but incorrect LLM-generated data are harmful as it trains retrievers with scientific falsehoods \cite{pal-etal-2023-med}.

\begin{comment}
0314 (예시만 바꿈-바꾸기 전 버전)
Specifically, scientific retrieval is governed by two structural properties: \textbf{long-tailed concept distribution} and \textbf{high fact-sensitivity}. 
Scientific corpora exhibit extreme long-tail distributions composed of irreplaceable entities such as chemical formulas and rare molecular structures \cite{oh-etal-2025-incorporating}. Unlike general-domain terms, these entities lack semantic substitutes, making naive data augmentation ineffective. As shown in Figure~\ref{fig:intro_overview}, LLMs often collapse distinct entities into scientifically invalid queries by conflating viruses and kinases.
Moreover, scientific outcomes are hypersensitive to precise experimental conditions. Misattributing pathway activation to the E7 rather than E6 oncoprotein can invert conclusions. Such plausible but incorrect LLM-generated data is particularly harmful because it forces retrievers to learn from scientific falsehoods \cite{pal-etal-2023-med}.

\end{comment}

\begin{comment}
    we pose a critical question: \textit{How can we effectively cover the extreme long-tail concepts of scientific domains while maintaining the rigorous factual precision?} In this paper, 
\end{comment}

To address these limitations, we propose \textbf{\textsc{REPAIR}} (\textbf{R}etriever via \textbf{Ep}istemic \textbf{A}PI-Guided \textbf{I}terative \textbf{R}efinement), an iterative, epistemic self-evolving procedure of data augmentation for training of an LLM-based scientific retriever. REPAIR first initializes a seed retriever with compact scientific corpora, then iteratively refines it through data synthesis of three stages: (1) \textbf{Diagnosis} identifies long-tail concepts that the retriever tends to confuse, (2) \textbf{Expansion} grounds these concepts in fact-verified documents mined from scientific APIs, and (3) \textbf{Differentiation} resolves fine-grained factual distinctions using fact-contrastive hard negatives to trains the model. Through iterative refinement, REPAIR progressively corrects the retriever's long-tail confusions using externally verified evidence.

We conduct comprehensive experiments across nine diverse scientific retrieval benchmarks in both materials science and biomedical domains. The results show that \textsc{REPAIR} enhances retrieval precision on specialized scientific tasks while exhibiting robust generalization capabilities. By iteratively correcting long-tail confusion with externally verified evidence, \textsc{REPAIR} outperforms 19 strong baselines, demonstrating the effectiveness of our iterative self-evolving strategy. In summary, this work presents the following contributions:

\begin{itemize}
\item We propose \textsc{REPAIR}, a self-evolving data augmentation framework that curtails structural hallucinations of scientific retrievers  by resolving long-tailed concept confusion and high fact-sensitivity, which have largely been overlooked in prior work.
\item Scaling retriever parameters from 500M to 7B, \textsc{REPAIR} achieves new state-of-the-art results across nine materials science and biomedical benchmarks, outperforming 19 strong baselines while using less training data.
\item Our extensive experiments show that the three-stage data augmentation pipeline of diagnosis, expansion, and differentiation outperforms naive synthetic data scaling in correcting long-tail retrieval errors.

\end{itemize}

\begin{comment}

\item We develop \textsc{REPAIR}, a scientific retriever scaling from 500M to 7B parameters, which mitigates LLM hallucination risks by explicitly resolving long-tailed concept confusion and high fact-sensitivity.
\item Through empirical analyses, we demonstrate that our API-guided factual verification mechanism successfully corrects persistent long-tail errors without relying on indiscriminate data scaling.
\item \textsc{REPAIR} achieves state-of-the-art performance across 9 scientific benchmarks, excelling in fundamental text representation tasks and retrieval-oriented applications specifically within materials science and biomedicine . It comprehensively outperforms 19 baselines with less training data [T].
\end{comment}

\section{Related Work}

A broad range of studies has investigated representation learning for text retrieval, progressing from latent semantic models to neural embedding-based approaches \cite{blei2003latent, hofmann1999probabilistic, deerwester1990indexing}. In recent years, dense retrieval with transformer encoders has become the dominant paradigm. Further gains have been achieved by scaling retrievers or applying instruction tuning with LLMs \cite{izacard2021unsupervised, yu2022coco, chen2024bge, wang2024improving, ni2022large, neelakantan2022text}. However, such improvements are largely attained in general-domain settings characterized by abundant supervised data.

As LLMs are increasingly applied to scientific reasoning, accurate retrieval of domain-specific knowledge has become critical for reliability \cite{zhang2025exploring, jiang2025applications, pilania2021machine, olivetti2020data}. Although retrievers have been adapted to scientific domains via domain-specific pretraining and task-oriented training \cite{jin2023medcpt, zhang2023pre, singh2023scirepeval}, scientific retrieval remains fundamentally challenged by distributional shifts \cite{kamalloo:2024}, particularly due to severe data scarcity and long-tailed entity distributions.

To address data scarcity, recent studies have adopted LLM-based data augmentation \cite{xu2024bmretriever}. However, such generative methods risk hallucinations, undermining the factual reliability essential for science \cite{pal-etal-2023-med}. Furthermore, while prior work on long-tailed distributions has focused on model-centric adaptations, such as specialized tokenization \cite{oh-etal-2025-incorporating} or domain-adaptive learning \cite{kim2024melt}, we argue that the fundamental bottleneck lies in the data themselves. Unlike previous model-centric strategies that attempt to adapt parameters to noisy or scarce distributions, our approach directly targets the quality and factual grounding of the retrieval data. We introduce a data-centric framework designed to mitigate the risks of hallucination and effectively cover long-tailed scientific concepts.

\begin{figure*}[t] % [t]는 페이지 상단에 배치하도록 지시
    \includegraphics[width=1\textwidth]{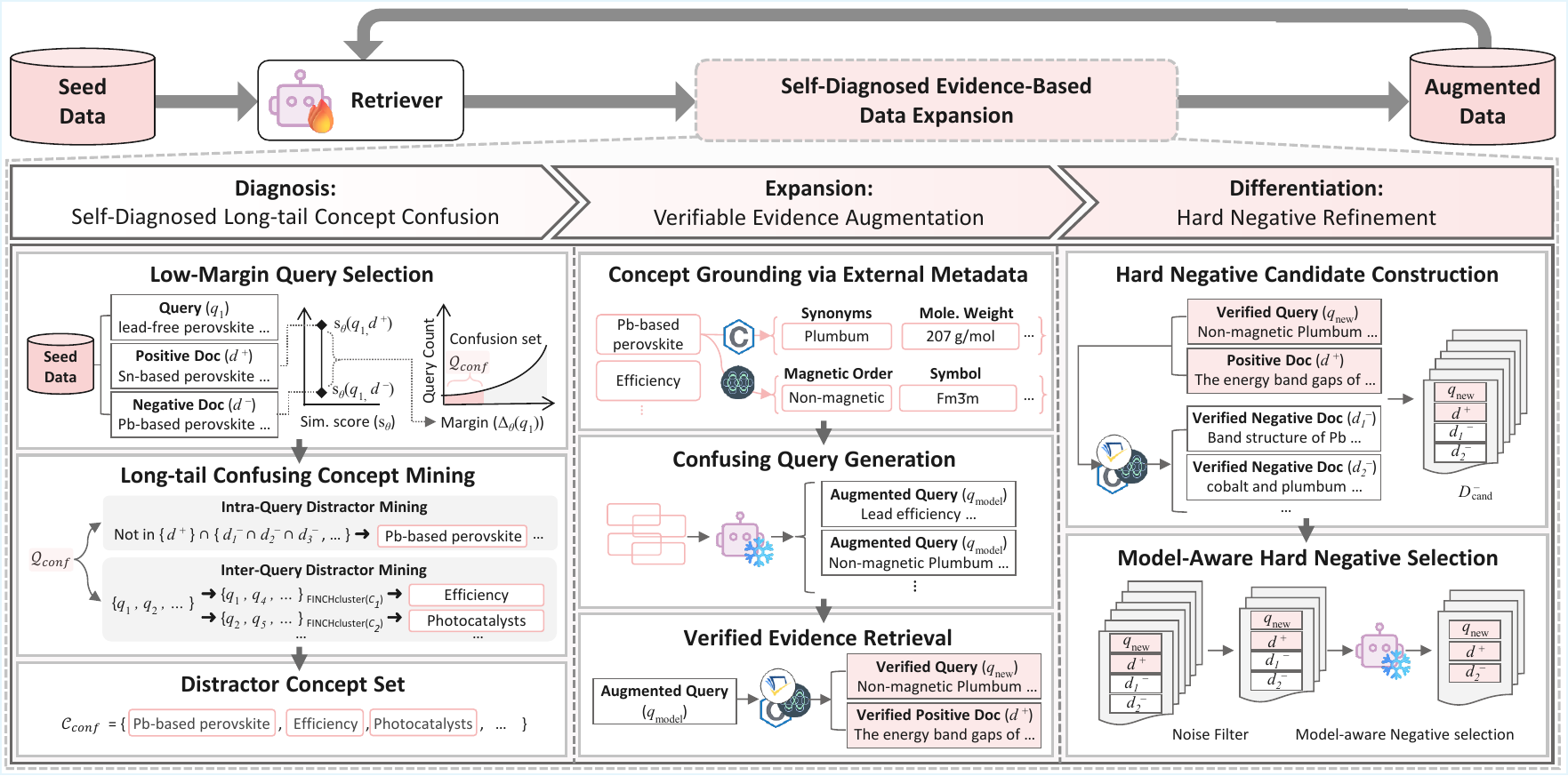} % 이미지 크기를 너비의 80%로 조정
    \caption{Overview of the \textsc{REPAIR} framework.
The model is initialized with scientific seed corpora and iteratively refined through self-diagnosed factual expansion, including diagnosis, expansion, and differentiation.} % 캡션
    \label{fig:main} % 참조를 위한 라벨
\end{figure*}

\section{Methodology: REPAIR}
\label{sec:method}

We focus on improving the reliability of dense retrievers for scientific domains (\S\ref{subsec:retriever}) by explicitly addressing long-tail concept confusion and high factual sensitivity. Starting from a retriever initialized with compact scientific seed corpora (\S\ref{subsec:init}), we refine it through three stages: \emph{Diagnosis} of long-tail concept uncertainty (\S\ref{subsec:diagnosis}), \emph{Expansion} with externally verified scientific evidence (\S\ref{subsec:expansion}), and \emph{Differentiation} via fact-contrastive hard negatives (\S\ref{subsec:diff}). This refinement is iteratively optimized using contrastive learning (\S\ref{subsec:train}). The overall procedure is illustrated in Figure \ref{fig:main}, and implementation details are provided in Appendix \ref{sec:implementation}.

% =========================================================

\subsection{Retriever Formulation}
\label{subsec:retriever}

Let $\mathcal{Q}$ be a set of queries and $\mathcal{D}$ a document corpus.
The dense retriever represents queries and documents as dense embeddings using a shared 
decoder-only language model $\mathcal{M}_\theta$ (e.g., Qwen-2.5 \cite{qwen2}) with parameter scales of 500M, 1.5B, and 7B.
For a query--document pair $(q,d)$, we compute the embeddings as
$\mathbf{e}_q \in \mathbb{R}^D$ and
$\mathbf{e}_d \in \mathbb{R}^D$; 
we append an end-of-sequence token to them
and compute their dense representations by EOS pooling over the final-layer hidden states:
\begin{equation}
\label{eq:eos_pool}
\mathbf{e}_{q/d} = \mathrm{Pool}_{\texttt{EOS}}\left(\mathcal{M}_\theta( q/d \oplus \texttt{[EOS]})\right).
\end{equation}
We score relevance by the dot product:
$s_\theta(q,d) = \mathbf{e}_q^\top \mathbf{e}_d$. 
For each query $q$, the retriever returns a ranked list
$\mathrm{TopK}_\theta(q) \subset \mathcal{D}$ under $s_\theta$.

% =========================================================
\subsection{Initialization from Scientific Seed Corpora}
\label{subsec:init}

\textsc{REPAIR} starts from a seed training set $\mathcal{T}_0=\{(q_i,d_i^+)\}_{i=1}^{N_0}$ constructed from well-recognized, domain-curated public scientific corpora. 
The full list is shown in Table \ref{tab:seed_data} with further details in Appendix \ref{Initial Corpus Construction}.
This seed set provides minimal in-domain alignment, but may be insufficient to cover long-tailed entities and condition-sensitive relations. 
Thus, \textsc{REPAIR} refines the retriever by iteratively incorporating verified supervision.

% =========================================================
\subsection{Stage I: Diagnosis}
\label{subsec:diagnosis}
In the first stage, the retriever self-diagnoses its weaknesses by identifying unreliable low-margin queries and extracts their long-tail concepts from its own scoring behavior.

\paragraph{Low-Margin Query Selection.}
%\label{subsec:lowmargin}
For each training query $q$ with a seed positive document $d^+(q)$,
we define the positive--negative separation margin:
\begin{equation}
\label{eq:margin}
\Delta_\theta(q)= s_\theta(q,d^+(q))  - \hspace{-6pt} \max_{d \in \mathrm{TopK}_\theta(q)\setminus\{d^+(q)\}} s_\theta(q,d).
\end{equation}
A small $\Delta_\theta(q)$ means that the retriever assigns nearly indistinguishable scores to $d^+(q)$ and top-ranked negatives, indicating local unreliability.
We form the confusion query set by selecting the lowest-margin queries:
\begin{equation}
\label{eq:confset}
\mathcal{Q}_{\mathrm{conf}}=\mathrm{Bottom}\text{-}p\% \big(\{\Delta_\theta(q)\}_{q\in\mathcal{Q}}\big),
\end{equation}
where we set $p=40\%$, whose empirical analysis is presented in \S\ref{ablation:iteration}.

\subsubsection*{Long-tail Confusing Concept Mining}
\label{subsec:conceptattrib}
While low margins reveal \emph{where} retrieval fails, REPAIR explains \emph{why} by identifying long-tail distractors driving model confusion. We first extract candidate concepts from $\mathcal{Q}_{\mathrm{conf}}$, and then isolate the actual distractor concepts via two complementary intra- and inter-query distractor mining.

\paragraph{Extraction of Candidate Concepts.}
For subsequent analysis, we extract candidate concepts $e \in \mathcal E$ such as scientific concepts and chemical formulas, by applying \textsc{MatDetector}~\cite{oh-etal-2025-incorporating} and \textsc{ChemDataExtractor}~\cite{swain2016chemdataextractor} to the confusing queries $q \in \mathcal{Q}_{\mathrm{conf}}$ and their retrieved documents.

\paragraph{Intra-Query Distractor Mining.}
To identify distractors specific to each query $q \in \mathcal{Q}_{\mathrm{conf}}$, we contrast the positive document $d^+(q)$ against a highly scored negative set $\mathcal{D}^-(q)$ as the top-$k$ retrieved documents. Aggregating over the confusion set,
\begin{equation}
\label{eq:posneg_aggr}
\mathcal{D}^+ = \{d^+(q)\}_{q\in\mathcal{Q}_{\mathrm{conf}}}, \,  
\mathcal{D}^- = \cup_{q\in\mathcal{Q}_{\mathrm{conf}}}\mathcal{D}^-(q), 
\end{equation}
we score each extracted concept $e$ using the Confusing Concept Score (CCS):
\begin{equation}
\label{eq:ccs_clean}
\mathrm{CCS}(e)=
\frac{\mathrm{df}(e;\mathcal{D}^-)}
{\mathrm{df}(e;\mathcal{D}^+)+\epsilon},
\end{equation}
where $\mathrm{df}(e;\cdot)$ is the document frequency of $e$, and $\epsilon>0$ prevents division by zero. Thus, a high CCS explicitly identifies distractor concepts that frequently occur in highly scored negative documents ($\mathcal{D}^-$) but remain rare in the positive documents ($\mathcal{D}^+$). Finally, we construct $\mathcal{C}_{\mathrm{intra}}$ by selecting the highest-CCS concept per query.

% -------------------------------
\paragraph{Inter-Query Distractor Mining.}
To complement the intra-query analysis, we identify systemic distractors by clustering queries within $\mathcal{Q}_{\mathrm{conf}}$ that exhibit shared confusion patterns, using the FINCH algorithm~\cite{sarfraz2019efficient}. Rather than relying on complex adjacency matrices, FINCH directly captures the mutual dependency between queries by grouping those that share first nearest neighbors. This parameter-free approach is critical for discovering confusion clusters without requiring a predefined cluster number. From each resulting cluster, we aggregate the previously extracted candidate concepts and select the most frequent one. This process transforms the clustered query groups into the inter-query concept set $\mathcal{C}_{\mathrm{inter}}$.

\paragraph{The Distractor Concept Set.} 
Finally, the distractor concept set is defined by $\mathcal{C}_{\mathrm{conf}} = \mathcal{C}_{\mathrm{intra}} \cup \mathcal{C}_{\mathrm{inter}}$, which identifies long-tail scientific concepts responsible for confusion. Then $\mathcal{C}_{\mathrm{conf}}$ is used in Stage~II for the expansion of verified evidence.

\subsection{Stage II: Expansion}
\label{subsec:expansion}
This stage expands the training data by grounding the distractor concept set $\mathcal{C}_{\mathrm{conf}}$ into verifiable evidence. This yields rigorously validated training tuples $(q_{\mathrm{new}}, d^+, \mathcal{D}^-_{\mathrm{cand}})$, consisting of a newly augmented query  $q_{\mathrm{new}}$, its positive document $d^+$, and its negative set $\mathcal{D}^-_{\mathrm{cand}}$.

\paragraph{Concept Grounding via External Metadata.}
We ground each concept in $\mathcal{C}_{\mathrm{conf}}$ using external databases such as \textsc{PubChem}~\cite{kim2019pubchem} and \textsc{MatProj}~\cite{jain2013commentary}, from which we extract diverse  chemical and physical attributes  of each concept (e.g., synonyms, molecular weight). %we construct an expanded representation for each concept.

\begin{comment}
  we ground each concept $e \in \mathcal{C}_{\mathrm{conf}}$ using external scientific databases. From \textsc{PubChem}, we extract \texttt{cmpdname}, \texttt{cmpdsynonym}, \texttt{mf}, \texttt{iupacname}, \texttt{meshheadings}, and \texttt{Molecular Weight}---a critical quantitative mass measure used to differentiate structurally similar compounds. From \textsc{MatProj}, we extract the chemical \texttt{Symbol} and \texttt{Magnetic Order}. These retrieved attributes are aggregated into a factual text description for each concept.
\end{comment}

\paragraph{Confusing Query Generation.}
From the grounded concepts, we randomly sample 1--3 concepts to prompt the model\footnote{Note that we use an LLM-based retriever (\S \ref{subsec:retriever}).}, which is instructed to generate a candidate query ($q_{\mathrm{model}}$) that it finds inherently ambiguous or difficult to resolve.

\paragraph{Verification and Hard Negative Mining.}
To filter out hallucinated $q_{\mathrm{model}}$, we query external APIs (\textsc{Semantic Scholar}, \textsc{PubChem}, and \textsc{MatProj}) and discard it if no results are returned. For valid searches, the top-matching document defines the ground truth: its title becomes the updated query $q_{\mathrm{new}}$, and its content serves as the positive document $d^+$. The remaining highly similar documents form $\mathcal{D}^-_{\mathrm{cand}}$ as hard negative candidates. In \S\ref{ablation:iteration}, we experiment with the effect of its size $|\mathcal{D}^-_{\mathrm{cand}}|$ on performance.
This generation and verification process continues until the number of valid tuples $(q_{\mathrm{new}}, d^+, \mathcal{D}^-_{\mathrm{cand}})$ reaches twice the size of the initial training data.

\subsection{Stage III: Differentiation}
\label{subsec:diff}
Instead of random in-batch negatives, we construct training triplets $(q_{\mathrm{new}}, d^+, d^-)$ by selecting a single hard negative $d^-$ from the API-verified candidates $\mathcal{D}^-_{\mathrm{cand}}$. 
Once mining noise is filtered out, we choose $d^-$ to confuse the current  retriever the most; $d^-$ is both factually plausible (API-ranked) and empirically challenging (model-scored).

\paragraph{Consistency Filtering.}
To prevent mining noise~\cite{wang2022text}, we define a localized candidate pool as $\mathcal{D}_{\mathrm{pool}} = \{d^+\} \cup \mathcal{D}^-_{\mathrm{cand}}$, and retain a tuple only if the current retriever ranks the positive document $d^+$ within the top $\kappa=2$ of this pool:
\begin{equation}
\label{eq:keep}
\mathbb{I}_{\mathrm{keep}}(q_{\mathrm{new}}) = \mathbb{1}\!\left[\mathrm{rank}\!\left(d^+ \mid q_{\mathrm{new}};\mathcal{D}_{\mathrm{pool}}\right) \le \kappa\right].
\end{equation}
This ensures the query is answerable, keeping the subsequent hard negative mining informative.

\paragraph{Single Hard Negative Selection.}
We extract the hardest negative $d^-$ from the candidate set $\mathcal{D}^-_{\mathrm{cand}}$ by maximizing the current retriever's similarity score $s_\theta$:
\begin{equation}
\label{eq:hardneg}
d^- = \operatorname*{argmax}_{d \in \mathcal{D}^-_{\mathrm{cand}}} s_\theta(q_{\mathrm{new}}, d).
\end{equation}

Using this single negative, we expect a more semantically meaningful decision boundary than when using multiple easy negatives. This stage completes a set of verified training triplets $(q_{\mathrm{new}}, d^+, d^-)$.

% =========================================================

\begin{comment}
    \subsection{Iterative Contrastive Optimization}
\label{subsec:train}

Let $\mathcal{N}(q_{\mathrm{new}})$ be a negative set containing $d^-(q_{\mathrm{new}})$ and in-batch negatives.
We optimize the parameter $\theta$ of the retriever with an InfoNCE objective:
\begin{equation}
\label{eq:infonce}
\mathcal{L}(q_{\mathrm{new}}) = -\log \frac{e^{s_\theta(q_{\mathrm{new}}, d^+)/\tau}}{\sum_{d \in \{d^+\} \cup \mathcal{N}} e^{s_\theta(q_{\mathrm{new}}, d)/\tau}}, 
\end{equation}
where $\tau$ is a temperature.
After updating $\theta$, the margin landscape $\{\Delta_\theta(q)\}$ changes; therefore \textsc{REPAIR} repeats Stages I--III.
This refinement progressively reshapes the embedding space toward reliable scientific discrimination, while avoiding hallucinated supervision.

[FB]
- 3.6에 iterative method 이면 iterative stopping condition을 쓰세요. 그냥 몇번 돌리고 마는건지 아니면 metric이 있는지

\end{comment}

\subsection{Iterative Contrastive Optimization}
\label{subsec:train}

The retriever parameters $\theta$ are updated via contrastive learning; we minimize an InfoNCE objective over the verified triplets $(q_{\mathrm{new}}, d^+, d^-)$:
\begin{equation}
\label{eq:infonce}
\mathcal{L}(q_{\mathrm{new}}) = -\log \frac{e^{s_\theta(q_{\mathrm{new}}, d^+)/\tau}}{\sum_{d \in \{d^+\} \cup \mathcal{N}} e^{s_\theta(q_{\mathrm{new}}, d)/\tau}}, 
\end{equation}
where $\tau$ is a temperature. As each update shifts the margin landscape $\{\Delta_\theta(q)\}$, \textsc{REPAIR} repeats the generation-verification pipeline (Stages I--III) for two iterations. We empirically study how performance varies with the number of iterations in \S\ref{sec:ablation}. This iterative refinement progressively reshapes the embedding space toward reliable scientific discrimination while avoiding hallucinated supervision or overfitting.

\section{Experiments}

% experiments + analysis 
\subsection{Experiment Setups}

\begin{comment}
 To assess the model's robustness, we utilize an extensive collection of datasets that cover a broad range of \textit{scientific} disciplines, from general inquiries to \textit{biomedical} and \textit{material}-specific challenges. The benchmark focuses on varied retrieval-oriented tasks, including four IR, four QA, one sentence similarity, one entity linking, and one paper recommendation task. Crucially, there is no overlap between the training and test pairs. Further details regarding the dataset statistics can be found in Appendix \ref{app:taskdataset}.
   
\end{comment}

\paragraph{Tasks and Datasets.} To assess the model's robustness, we use an extensive collection of benchmarks that cover a broad range of \textit{scientific} disciplines, from general inquiries to \textit{material} and \textit{biomedical}-specific challenges. They evaluate varied retrieval-oriented tasks, including four IR datasets (NFCorpus \cite{boteva2016full}, SciFact \cite{wadden2020fact}, SciDocs \cite{cohan2020specter}, and TREC-COVID \cite{voorhees2021trec}), three QA datasets (iCliniq \cite{DBLP:journals/corr/abs-2004-03329}, and the materials and biomedical subsets of ChemLit-QA \cite{DBLP:journals/mlst/WellawatteGLBHBS25}), one entity linking (MeSH \cite{lipscomb2000medical}), one paper recommendation (RELISH \cite{singh2023scirepeval, DBLP:journals/biodb/BrownCZ19}), and one sentence similarity dataset (BIOSSES \cite{souganciouglu2017biosses}). Full details about datasets are provided in Appendix \ref{app:taskdataset}.

\paragraph{Baselines.}
We compare our method with an extensive set of 19 baselines.
They include 
one sparse retriever such as BM25~\cite{robertson2009probabilistic} and 14 dense retrievers across various model scales, such as Contriever~\cite{izacardunsupervised}, Dragon~\cite{DBLP:conf/emnlp/LinALOLMY023}, InstructOR-L/XL~\cite{su2023one}, E5-Large-v2~\cite{wang2022text}, BGE-Large~\cite{chen2024m3}, DRAMA-L/1B~\cite{ma2025drama}, GTR-XL/XXL~\cite{ni2022large}, SGPT-1.3B/2.7B~\cite{muennighoff2022sgpt} and Llama2Vec~\cite{li2024llama2vec}, RepLLaMA~\cite{ma2024fine}, LLM2Vec~\cite{behnamghader2024llm2vec}, E5-Mistral~\cite{wang2024improving}, CPT-text-XL~\cite{neelakantan2022text}, and Promptriever~\cite{DBLP:conf/iclr/WellerDLPZH25}. We also include four models specialized for scientific domains:  SciMult~\cite{zhang2023pre}, SPECTER 2.0~\cite{singh2023scirepeval}, MedCPT~\cite{jin2023medcpt}, and \textsc{BMRetriever} series (410M/2B/7B)~\cite{xu2024bmretriever}. Details about the baselines are provided in the Appendix \ref{app:baselines}.

\paragraph{Training.}
 We train Qwen2.5-0.5B/1.5B/7B with scientific seed data with a particular focus on materials science \cite{tshitoyan2019unsupervised, gupta2022matscibert, trewartha2022quantifying} and biomedical domains \cite{bajaj2016ms, wang-etal-2020-cord, xiong2024benchmarking, chen2021litcovid}. More training details  are provided in Appendix \ref{app:implementation}.

\begin{table*}[!t]
\centering
\arraybackslash
\fontsize{4}{7}\selectfont % 9pt 글씨, 11pt 줄 간격
\setlength{\tabcolsep}{4pt} % 기본값은 6pt
\renewcommand{\arraystretch}{0.8} % 행 높이를 1.5배로 늘림
\resizebox{\textwidth}{!}{%
\begin{tabular}{lclcccccccc}
\hline
\multicolumn{1}{l|}{Task} & \multirow{2}{*}{Scale} & \multicolumn{1}{c}{\multirow{2}{*}{\# Pairs}} & \multicolumn{1}{c|}{\multirow{2}{*}{\begin{tabular}[c]{@{}c@{}}Data\\ Aug.\end{tabular}}} & \multicolumn{4}{c|}{Standard IR} & \multicolumn{1}{c|}{\multirow{2}{*}{AVG.}} & \multicolumn{1}{c|}{Sent. Sim.} & \multirow{2}{*}{AVG.} \\ \cline{1-1} \cline{5-8} \cline{10-10}
\multicolumn{1}{l|}{Model} &  & \multicolumn{1}{c}{} & \multicolumn{1}{c|}{} & NFCorpus & SciFact & SciDocs & \multicolumn{1}{c|}{COVID} & \multicolumn{1}{c|}{} & \multicolumn{1}{c|}{BIOSSES} &  \\ \hline

\multicolumn{1}{l|}{BM25} & - & \multicolumn{1}{c}{-} &\multicolumn{1}{c|}{\cmark}& 0.325 & 0.665 & 0.158 & \multicolumn{1}{c|}{0.656} & \multicolumn{1}{c|}{0.451} & \multicolumn{1}{c|}{-} &  \\ \hline

\multicolumn{1}{l|}{Contriever} & 110M & \multicolumn{1}{c}{1.5B} &\multicolumn{1}{c|}{\cmark} & 0.328 & 0.677 & 0.165 & \multicolumn{1}{c|}{0.596} & \multicolumn{1}{c|}{0.442} & \multicolumn{1}{c|}{0.833} & 0.520 \\

\multicolumn{1}{l|}{Dragon} & 110M & \multicolumn{1}{c}{28.5M} &\multicolumn{1}{c|}{\cmark} & 0.339 & 0.679 & 0.159 & \multicolumn{1}{c|}{0.759} & \multicolumn{1}{c|}{0.484} & \multicolumn{1}{c|}{0.819} & 0.540 \\

 \rowcolor[HTML]{   EBF2F6}\multicolumn{1}{l|}{SPECTER 2.0} & 110M & \multicolumn{1}{c}{3.3M} &\multicolumn{1}{c|}{}  & 0.228 & 0.671 & - & \multicolumn{1}{c|}{0.584} & \multicolumn{1}{c|}{-} & \multicolumn{1}{c|}{-} & \multicolumn{1}{c}{-} \\
\rowcolor[HTML]{   EBF2F6}\multicolumn{1}{l|}{SciMult} & 110M & \multicolumn{1}{c}{5.5M} &\multicolumn{1}{c|}{}  & 0.308 & 0.707 & - & \multicolumn{1}{c|}{0.712} & \multicolumn{1}{c|}{-} & \multicolumn{1}{c|}{-} & \multicolumn{1}{c}{-} \\
\rowcolor[HTML]{   EBF2F6}\multicolumn{1}{l|}{MedCPT} & 220M & \multicolumn{1}{c}{255M} &\multicolumn{1}{c|}{\cmark}  & 0.340 & 0.724 & 0.123 & \multicolumn{1}{c|}{0.697} & \multicolumn{1}{c|}{0.471} & \multicolumn{1}{c|}{0.837} & 0.532 \\

\multicolumn{1}{l|}{InstructOR-L} & 335M & \multicolumn{1}{c}{1.24M} &\multicolumn{1}{c|}{\cmark}  & 0.341 & 0.643 & 0.186 & \multicolumn{1}{c|}{0.581} & \multicolumn{1}{c|}{0.438} & \multicolumn{1}{c|}{0.844} & 0.505 \\
\multicolumn{1}{l|}{E5-Large-v2†} & 660M & \multicolumn{1}{c}{271M} &\multicolumn{1}{c|}{\cmark}  & 0.371 & 0.726 & 0.201 & \multicolumn{1}{c|}{0.665} & \multicolumn{1}{c|}{0.491} & \multicolumn{1}{c|}{0.836} & 0.548 \\

\multicolumn{1}{l|}{BGE-Large∗‡} & 895M & \multicolumn{1}{c}{2.8B} &\multicolumn{1}{c|}{\cmark}  & 0.345 & 0.723 & 0.222 & \multicolumn{1}{c|}{0.753} & \multicolumn{1}{c|}{\underline{0.511}} & \multicolumn{1}{c|}{0.804} & 0.560 \\
\rowcolor[HTML]{   EBF2F6}\multicolumn{1}{l|}{\textsc{BMRetriever}-410M} & 410M & \multicolumn{1}{c}{11.4M} &\multicolumn{1}{c|}{\cmark}  & 0.321 & 0.711 & 0.167 & \multicolumn{1}{c|}{0.831} & \multicolumn{1}{c|}{0.508} & \multicolumn{1}{c|}{0.840} & \underline{0.563} \\
\multicolumn{1}{l|}{DRAMA-L} & 300M & \multicolumn{1}{c}{127M} &\multicolumn{1}{c|}{\cmark}  & 0.324 & 0.651 & 0.138 & \multicolumn{1}{c|}{0.500} & \multicolumn{1}{c|}{0.403} & \multicolumn{1}{c|}{0.725} & 0.442 \\
\rowcolor[HTML]{   EBF2F6} 
\multicolumn{1}{l|}{\cellcolor[HTML]{   EBF2F6}\textsc{REPAIR-500M} (ours)} & 500M & \multicolumn{1}{c}{\cellcolor[HTML]{   EBF2F6}4M} &\multicolumn{1}{c|}{\cellcolor[HTML]{   EBF2F6}\cmark}  & 0.376 & 0.680 & 0.196 & \multicolumn{1}{c|}{\cellcolor[HTML]{   EBF2F6}0.812} & \multicolumn{1}{c|}{\cellcolor[HTML]{   EBF2F6}\textbf{0.516}} & \multicolumn{1}{c|}{\cellcolor[HTML]{   EBF2F6}0.853} & \textbf{0.583} \\ \hline

\multicolumn{1}{l|}{InstructOR-XL} & 1.5B & \multicolumn{1}{c}{1.24M}&\multicolumn{1}{c|}{\cmark} & 0.360 & 0.646 & 0.174 & \multicolumn{1}{c|}{0.713} & \multicolumn{1}{c|}{0.473} & \multicolumn{1}{c|}{0.842} & 0.547 \\
\multicolumn{1}{l|}{GTR-XL} & 1.2B & \multicolumn{1}{c}{2.7B} &\multicolumn{1}{c|}{\cmark} & 0.343 & 0.635 & 0.159 & \multicolumn{1}{c|}{0.584} & \multicolumn{1}{c|}{0.430} & \multicolumn{1}{c|}{0.789} & 0.502 \\
\multicolumn{1}{l|}{GTR-XXL} & 4.8B & \multicolumn{1}{c}{2.7B} &\multicolumn{1}{c|}{\cmark} & 0.342 & 0.662 & 0.161 & \multicolumn{1}{c|}{0.501} & \multicolumn{1}{c|}{0.417} & \multicolumn{1}{c|}{0.819} & 0.497 \\
\multicolumn{1}{l|}{SGPT-1.3B} & 1.3B & \multicolumn{1}{c}{unknown} &\multicolumn{1}{c|}{\cmark} & 0.320 & 0.682 & 0.162 & \multicolumn{1}{c|}{0.730} & \multicolumn{1}{c|}{0.473} & \multicolumn{1}{c|}{0.830} & 0.545 \\
\multicolumn{1}{l|}{SGPT-2.7B} & 2.7B & \multicolumn{1}{c}{unknown} &\multicolumn{1}{c|}{\cmark} & 0.339 & 0.701 & 0.166 & \multicolumn{1}{c|}{0.752} & \multicolumn{1}{c|}{0.489} & \multicolumn{1}{c|}{0.848} & 0.561 \\
\rowcolor[HTML]{   EBF2F6}\multicolumn{1}{l|}{\textsc{BMRetriever}-2B} & 2B & \multicolumn{1}{c}{10M} &\multicolumn{1}{c|}{\cmark} & 0.351 & 0.760 & 0.199 & \multicolumn{1}{c|}{0.863} & \multicolumn{1}{c|}{\underline{0.543}} & \multicolumn{1}{c|}{0.828} & \underline{0.600} \\
\multicolumn{1}{l|}{DRAMA-1B} & 1B & \multicolumn{1}{c}{127M} &\multicolumn{1}{c|}{\cmark}& 0.158 & 0.707 & 0.145 & \multicolumn{1}{c|}{0.412} & \multicolumn{1}{c|}{0.355} & \multicolumn{1}{c|}{0.765} & 0.419 \\
\rowcolor[HTML]{   EBF2F6} 
\multicolumn{1}{l|}{\cellcolor[HTML]{   EBF2F6}\textsc{REPAIR-1.5B} (ours)} & 1.5B & \multicolumn{1}{c}{\cellcolor[HTML]{   EBF2F6}4M} &\multicolumn{1}{c|}{\cmark} & 0.376 & 0.757 & 0.201 & \multicolumn{1}{c|}{\cellcolor[HTML]{   EBF2F6}0.853} & \multicolumn{1}{c|}{\cellcolor[HTML]{   EBF2F6}\textbf{0.546}} & \multicolumn{1}{c|}{\cellcolor[HTML]{   EBF2F6}0.849} & \textbf{0.607} \\ \hline

\multicolumn{1}{l|}{Llama2Vec} & 7B & \multicolumn{1}{c}{21.5M} &\multicolumn{1}{c|}{\cmark} & 0.372 & 0.757 & 0.172 & \multicolumn{1}{c|}{0.853} & \multicolumn{1}{c|}{0.539} & \multicolumn{1}{c|}{-} & - \\
\multicolumn{1}{l|}{RepLLaMA} & 7B & \multicolumn{1}{c}{500K} &\multicolumn{1}{c|}{\cmark} & 0.378 & 0.756 & 0.181 & \multicolumn{1}{c|}{0.847} & \multicolumn{1}{c|}{0.541} & \multicolumn{1}{c|}{-} & - \\
\multicolumn{1}{l|}{LLM2Vec} & 7B & \multicolumn{1}{c}{2.7M} &\multicolumn{1}{c|}{\cmark} & 0.393 & 0.788 & 0.225 & \multicolumn{1}{c|}{0.776} & \multicolumn{1}{c|}{0.545} & \multicolumn{1}{c|}{0.852} & 0.606 \\
\multicolumn{1}{l|}{E5-Mistral} & 7B & \multicolumn{1}{c}{1.8M}&\multicolumn{1}{c|}{\cmark} & 0.386 & 0.764 & 0.162 & \multicolumn{1}{c|}{0.872} & \multicolumn{1}{c|}{0.546} & \multicolumn{1}{c|}{0.855} & 0.608 \\
\multicolumn{1}{l|}{CPT-text-XL} & 175B & \multicolumn{1}{c}{unknown} &\multicolumn{1}{c|}{}& 0.407 & 0.754 & - & \multicolumn{1}{c|}{0.649} & \multicolumn{1}{c|}{-} & \multicolumn{1}{c|}{-} & - \\
\rowcolor[HTML]{   EBF2F6}\multicolumn{1}{l|}{\textsc{BMRetriever}-7B} & 7B & \multicolumn{1}{c}{11.4M} &\multicolumn{1}{c|}{\cmark}& 0.364 & 0.778 & 0.201 & \multicolumn{1}{c|}{0.861} & \multicolumn{1}{c|}{\underline{0.551}} & \multicolumn{1}{c|}{0.847} & \underline{0.610} \\
\multicolumn{1}{l|}{Promptriever} & 7B & \multicolumn{1}{c}{1M} &\multicolumn{1}{c|}{\cmark} & 0.376 & 0.760 & 0.176 & \multicolumn{1}{c|}{0.835} & \multicolumn{1}{c|}{0.537} & \multicolumn{1}{c|}{0.861} & 0.602 \\
\rowcolor[HTML]{   EBF2F6} 
\multicolumn{1}{l|}{\cellcolor[HTML]{   EBF2F6}\textsc{REPAIR-7B} (ours)} & 7B & \multicolumn{1}{c}{\cellcolor[HTML]{   EBF2F6}4M}&\multicolumn{1}{c|}{\cmark} & 0.413 & 0.789 & 0.227 & \multicolumn{1}{c|}{\cellcolor[HTML]{   EBF2F6}0.842} & \multicolumn{1}{c|}{\cellcolor[HTML]{   EBF2F6}\textbf{0.568}} & \multicolumn{1}{c|}{\cellcolor[HTML]{   EBF2F6}0.846} & \textbf{0.623} \\ \hline

\end{tabular}
}

\caption{Experiments on scientific text representation tasks across various model scales. All scores are reported in nDCG@10. $\dagger$ and $\ddagger$ denote the use of reranker distillation and hybrid retrieval, respectively. We highlight the \colorbox[HTML]{   EBF2F6}{scientific} domain-specific retrieval models. "\#Pairs" and "Sent. Sim." stand for the total number of query-document pairs used for training and Sentence Similarity, respectively. The best-performing results are highlighted in \textbf{boldface}, while \underline{underlined}  represent the second-highest scores.}

  \label{tab:mainresult}
\end{table*}

\begin{table*}[ht]
\centering
\arraybackslash
\fontsize{6.3}{6}\selectfont % 9pt 글씨, 11pt 줄 간격
\setlength{\tabcolsep}{3pt} % 기본값은 6pt
\renewcommand{\arraystretch}{1.6} % 행 높이를 1.5배로 늘림
\resizebox{\textwidth}{!}{%
\newcommand{\w}[1]{\makebox[\widthof{nDCG@20}]{#1}}
\begin{tabular}{l|ccc|ccc|ccc|ccc|ccc|cc}

\hline
\multicolumn{4}{l|}{Task} & \multicolumn{9}{c|}{Question Answering} & \multicolumn{3}{c|}{Entity Linking} & \multicolumn{2}{c}{Paper Rec.} \\ \hline

\multirow{2}{*}{Model} & \multirow{2}{*}{Scale} & \multirow{2}{*}{\# Pairs} & \multirow{2}{*}{\begin{tabular}[c]{@{}c@{}}Data\\      Aug.\end{tabular}} & \multicolumn{3}{c|}{iCliniq} & \multicolumn{3}{c|}{ChemLit-QA$_{\text{mat}}$} & \multicolumn{3}{c|}{ChemLit-QA$_{\text{biomed}}$} & \multicolumn{3}{c|}{MeSH} & \multicolumn{2}{c}{RELISH} \\ \cline{5-18} 
 &  &  &  & R@5 & R@20 & nDCG & R@5 & R@20 & nDCG & R@5 & R@20 & nDCG & R@1 & R@5 & MRR@5 & MAP & nDCG \\ \hline
Dragon & 110M & 28.5M & \cmark & 50.6 & 65.2 & 47.4 & 70.9 & 92.7 & 60.5 & 70.3 & 94.3 & 63.8 & 28.2 & 47.0 & 34.8 & 72.6 & 80.6 \\
\rowcolor[HTML]{   EBF2F6}MedCPT & 220M & 255M & \cmark & 26.8 & 42.0 & 24.9 & 61.3 & 89.5 & 53.7 & 66.1 & 90.3 & 64.4 & 27.7 & 54.2 & 37.4 & 83.6 & 89.7 \\
E5-Large-v2† & 660M & 271M & \cmark & 57.6 & 72.0 & 55.8 & \underline{74.3} & 92.2 & 61.3 & \underline{79.8} & \underline{94.5} & \underline{65.1} & \underline{32.8} & \underline{55.0} & \underline{41.3} & 84.9 & 91.0 \\
\rowcolor[HTML]{   EBF2F6}\textsc{BMRetriever}-410M & 410M & 11.4M & \cmark & \underline{60.6} & \underline{72.8} & \underline{56.6} & 72.7 & \underline{92.1} & \underline{64.0} & 72.7 & 94.0 & 64.0 & 31.5 & 53.8 & 39.8 & \underline{85.2} & \underline{91.2} \\
\rowcolor[HTML]{   EBF2F6} \textsc{REPAIR-500M}   (ours) & 500M & 4M & \cmark & \textbf{61.3} & \textbf{74.1} & \textbf{57.2} & \textbf{75.2} & \textbf{92.8} & \textbf{64.4} & \textbf{81.0} & \textbf{94.7} & \textbf{65.2} & \textbf{33.8} & \textbf{57.7} & \textbf{42.8} & \textbf{85.8 }& \textbf{91.5} \\ \hline

InstructOR-XL & 1.5B & 1.24M & \cmark & 64.9 & 78.1 & 58.3 & 64.0 & 85.4 & 51.9 & 74.6 & 92.0 & 60.1 & 33.6 & 56.2 & 45.7 & 84.5 & 90.6 \\
SGPT-2.7B & 2.7B & unknown & \cmark & 45.0 & 52.2 & 41.2 & 56.8 & 90.1 & 49.6 & 70.9 & 89.4 & 56.0 & 33.6 & 56.2 & 45.7 & 84.5 & 90.6 \\
\rowcolor[HTML]{   EBF2F6}\textsc{BMRetriever}-2B & 2B & 10M & \cmark & \textbf{70.0} & \textbf{81.2} & \textbf{65.7} & \underline{83.1} & \underline{93.1} & \underline{61.4} & \underline{87.7} & \underline{96.3} & \underline{72.6} & \textbf{45.6} & \textbf{71.3} & \textbf{59.5} & \underline{85.4} & \textbf{91.5} \\
\rowcolor[HTML]{   EBF2F6} \textsc{REPAIR-1.5B} (ours) & 1.5B & 4M & \cmark & \underline{65.0} & \underline{80.8} & \underline{61.7} & \textbf{83.7} & \textbf{96.0} & \textbf{68.3} & \textbf{88.7} & \textbf{97.4} & \textbf{72.9} & \underline{41.2} & \underline{70.5} & \underline{51.7} & \textbf{85.7} & \underline{90.9} \\ \hline

E5-Mistral & 7B & 1.8M & \cmark & 56.7 & 72.2 & 51.8 & 80.9 & 94.0 & 64.0 & \underline{86.7} & \underline{96.4} & 70.1 & 47.9 & 76.2 & \underline{61.3} & 85.2 & 90.8 \\
\rowcolor[HTML]{   EBF2F6}\textsc{BMRetriever}-7B & 7B & 11.4M & \cmark & \underline{68.4} & \underline{79.7} & \underline{63.7} & \underline{82.2} & \underline{94.1} & \underline{67.2} & 86.2 & 95.4 & \underline{71.1} & \underline{49.8} & \underline{76.5} & 61.1 & \underline{86.7} & \underline{92.2} \\ 
\rowcolor[HTML]{   EBF2F6} \textsc{REPAIR-7B} (ours) & 7B & 4M & \cmark & \textbf{70.1} & \textbf{81.3} & \textbf{65.6} & \textbf{86.0} & \textbf{97.8} & \textbf{71.5} & \textbf{91.5} & \textbf{99.4} & \textbf{73.5} & \textbf{51.0} & \textbf{78.2} & \textbf{62.0} & \textbf{87.4} & \textbf{92.8}\\\hline
\end{tabular}
}
    \caption{Experiments on retrieval-oriented material and biomedical NLP applications  across materials science and biomedical domains. Here, $\text{ChemLit-QA}_{\text{mat}}$ and $\text{ChemLit-QA}_{\text{biomed}}$ denote the materials and biomedical categories of the ChemLit-QA task, respectively. nDCG refers to nDCG@20, except for the paper recommendation task. The best-performing results are highlighted in \textbf{boldface}, while \underline{underline} represent the second-highest scores.}

  \label{tab:maintable2}
\end{table*}

\paragraph{Evaluation.}

\begin{comment}
    We evaluate retrieval effectiveness using task-specific metrics. Following~\citet{xu2024bmretriever}, standard information retrieval is evaluated with nDCG@10, and sentence similarity with Spearman’s rank correlation over cosine similarity. For question answering, we report Recall@5, Recall@20, and nDCG@20, while entity linking is evaluated using MRR@5 and Recall@1/5. For paper recommendation, we follow~\cite{singh2023scirepeval} and report MAP and nDCG.

    We evaluate retrieval accuracy using task-specific metrics, following~\citet{xu2024bmretriever}. Standard information retrieval is evaluated with nDCG@10, and sentence similarity with Spearman’s rank correlation over cosine similarity. For question answering, we report Recall@5, Recall@20, and nDCG@20.

\end{comment}

To ensure rigorous evaluation, we follow all experiment setups of BMRetriever \citep {xu2024bmretriever}, including dataset curation, task formulation, baseline selection, and evaluation metrics. Following this framework, we categorize our evaluation into two distinct areas: fundamental text representation tasks (Table \ref{tab:mainresult}) and retrieval-oriented material and biomedical applications (Table \ref{tab:maintable2}). Standard information retrieval is evaluated with nDCG@10, and sentence similarity with Spearman's rank correlation over cosine similarity. For material and biomedical applications, we report Recall@\{5, 20\} and nDCG@20 for question answering, mean reciprocal rank (MRR)@5 and Recall@\{1, 5\} for entity linking, and mean average precision (MAP) and nDCG for paper recommendation \citep{singh2023scirepeval}.

\begin{comment}
    8:1:1

    Split results - Train: 2559490, Valid: 319936, Test: 319937
\end{comment}

\subsection{Main Results}

\paragraph{Results on Text Representation Tasks.}
Table~\ref{tab:mainresult} presents a comprehensive evaluation of embedding quality across four science IR and one sentence similarity benchmarks. Across different scales, \textsc{REPAIR} consistently outperforms baseline methods. While some strong baselines heavily rely on computationally expensive reranker distillation (e.g., E5-Large-v2$^\dagger$~\cite{wang2022text}) or complex hybrid systems requiring sparse inverted indices (e.g., BGE-Large$^\ddagger$~\cite{chen2024m3}), %\textsc{REPAIR} achieves superior performance as a purely standalone dense retriever without any external dependency or additional supervision signals. 

\textsc{REPAIR} demonstrates exceptional parameter and data efficiency. First, in terms of parameter efficiency, it exhibits  competitive performance against substantially larger baselines. Specifically, \textsc{REPAIR}-500M successfully surpasses both the SGPT-2.7B~\cite{muennighoff2022sgpt} and the GTR-XXL~\cite{ni2022large} with 4.8B parameters. Furthermore, \textsc{REPAIR}-1.5B outperforms massive 7B LLM-based retrievers, such as LLM2Vec~\cite{behnamghader2024llm2vec} and Promptriever~\cite{DBLP:conf/iclr/WellerDLPZH25}. Second, from a data efficiency perspective, \textsc{REPAIR} uses only 4M fact-verified instances. In stark contrast, it significantly exceeds the performance of \textsc{BMRetriever}-2B, which consumes 11.4M synthetic pairs, as well as BGE-Large, a model trained on an extensive corpus of 2.8B text pairs. %Finally, the efficacy of this fact-driven grounding is most pronounced in domains demanding rigorous precision. On the SciFact benchmark, \textsc{Repair}-1.5B secures an outstanding score of 0.781, outperforming \textsc{BMRetriever}-2B (0.701) by a substantial margin of +0.080. Similarly, it achieves a peak performance of 0.867 on the TREC-COVID dataset, underscoring its robustness in stringent scientific retrieval tasks.

\begin{comment}
    As detailed in Table \ref{tab:mainresult}, REPAIR consistently achieves strong performance on average across benchmarks, explicitly demonstrating that targeted factual grounding of long-tail concepts is more critical than mere data scaling. While existing methods like BMRETRIEVER~\cite{xu2024bmretriever} rely on massive, hallucination-prone synthetic datasets, \textsc{REPAIR} consistently outperforms these data-intensive approaches using only a 4M rigorously verified training set. This indicates that diagnosing epistemic uncertainty to curate fact-based evidence effectively resolves fine-grained distinctions that even larger-scale generative augmentation fails to capture. Notably, \textsc{REPAIR} surpasses models utilizing complex reranker distillation \cite{wang2022text} or hybrid retrieval \cite{chen2024m3}. This suggests that our iterative refinement progressively reshapes the embedding space into a more discriminative manifold, achieving high-precision retrieval without the computational overhead of multi-stage architectures.

\end{comment}

\paragraph{Results on Retrieval-Oriented Material and Biomedical Applications.}

\begin{comment}
    Table \ref{tab:maintable2} highlights \textsc{REPAIR}’s robust generalization across specialized downstream tasks. While \textsc{BMRetriever-2B}~\cite{xu2024bmretriever} exhibits slightly higher performance in the biomedical domain, this is largely expected as it relies heavily on its larger scale (2B) and an exhaustive multi-task instruction fine-tuning stage. Specifically, it necessitates gathering human-annotated datasets and generating additional synthetic retrieval tasks under various scenarios to augment training samples and diversify instructions. In contrast, \textsc{REPAIR-1.5B} utilizes a unified, task-agnostic query format, eliminating the reliance on curating such a diverse collection of task-specific query-passage pairs. Despite this highly streamlined approach to data construction, \textsc{REPAIR} remains incredibly competitive. It seamlessly adapts to diverse scenarios—from question answering to entity linking—demonstrating exceptional parameter efficiency and outperforming specialized models like MedCPT~\cite{jin2023medcpt}. By achieving these results without the massive overhead of compiling diverse retrieval tasks into the instruction tuning blend, \textsc{REPAIR} proves that fact-verified grounding establishes a universally adaptable semantic space.

\end{comment}

\begin{figure}[t] % Use [H] to force the figure to stay here
    \centering % Centers the figure
    \includegraphics[width=\columnwidth]{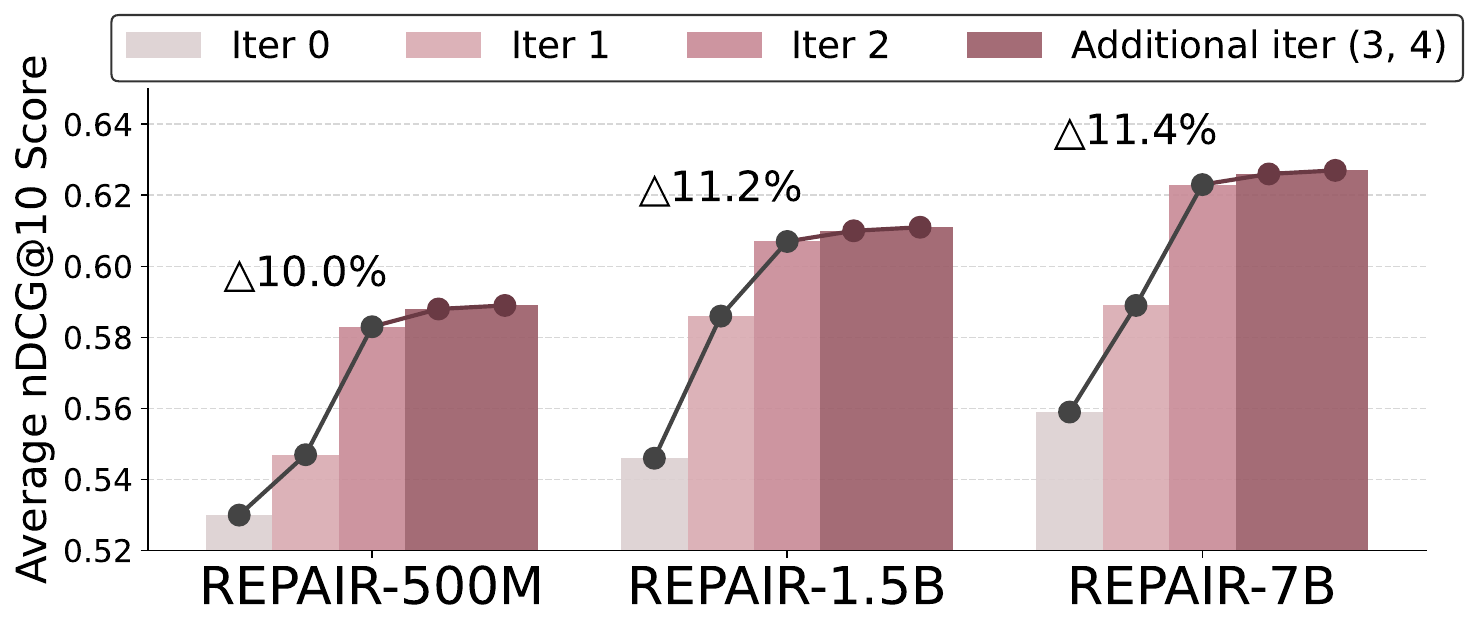} % 이미지 크기를 너비의 50%로 조정
    \caption{Performance variations of \textsc{REPAIR} models according to the number of iterations. Iteration 0 indicates the seed-only training baseline. The percentages above the bars are the relative nDCG@10 improvement of Iteration 2 over Iteration 0.} % 캡션
    \label{fig:abb_iter} % 참조를 위한 라벨
\end{figure}

\begin{figure*}[t] % [t]는 페이지 상단에 배치하도록 지시
    \includegraphics[width=1\textwidth]{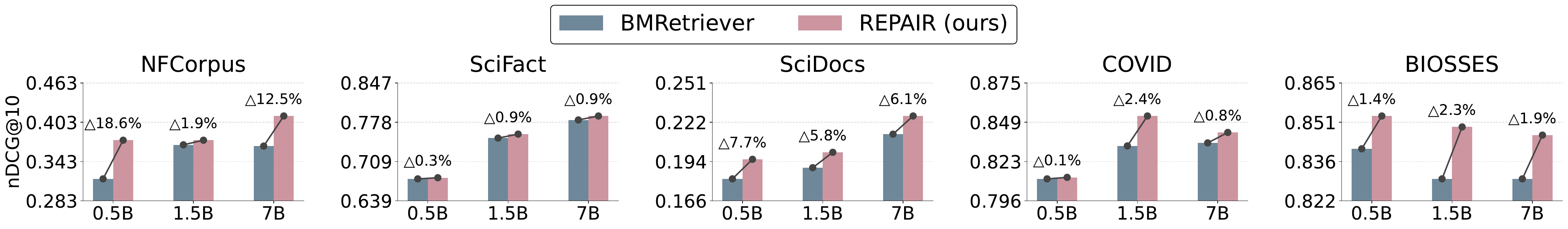} % 이미지 크기를 너비의 80%로 조정
    \caption{Effect of fact-verified data across model capacities. The evaluation is based on the nDCG@10 metric using three different model sizes (0.5B, 1.5B, and 7B).} % 캡션
    \label{fig:qwen} % 참조를 위한 라벨
\end{figure*}

\begin{table*}[htbp]
\small
\renewcommand{\arraystretch}{1.1}
% 표 양끝에 1em 크기의 정밀한 여백을 추가했습니다.
\begin{tabularx}{\textwidth}{@{\hspace{0.5em}} l X @{\hspace{0.5em}}}
\hline
\multicolumn{2}{c}{Domain: Material \quad | \quad Concept: Pb-based perovskite \quad | \quad Attribute: Plumbum, Non-magnetic, \dots} \\
\hline
Query ($q_\text{new}$) & Dynamic symmetry breaking and spin splitting in metal halide \myhl{{perovskites}}. \\
Positive ($d^{+}$) & While materials such as $\mathrm{CH_3NH_3PbI_3}$ are \myhl{{nonmagnetic}}, the presence of heavy elements (\myhl{{Pb}} and I) in a noncentrosymmetric crystal environment result in a spin splitting of the frontier electronic bands \dots \\
Negative ($d^{-}$) & We report a theoretical investigation of Rashba band splitting in ferroelectric halide \myhl{{perovskite materials}}. Since the polarization direction in ferroelectric materials can be switched by external electric fields \dots \\
\hline
\multicolumn{2}{c}{Domain: Biomedical \quad | \quad Concept: lactic acid \quad | \quad Attribute: DL-Lactic acid, $\mathrm{C_3H_6O_3}$ \dots} \\
\hline
Query ($q_\text{new}$) & Hydrolytic degradation of devices based on poly(\myhl{dl-lactic acid}) size-dependence. \\
Positive ($d^{+}$) & Millimetric beads and submillimetric microspheres and cast films, derived from the same batch of poly (\myhl{dl-lactic acid}) polymer were allowed to age comparatively in isoosmolar 0.13 m phosphate \dots \\
Negative ($d^{-}$) & To improve the reversibility of zinc plating/stripping for high-performance AZIBs. \myhl{C$_3$H$_6$O$_3$} with high zinc adsorption ability not only can reconstruct the Zn$^{2+}$ solvation sheath, reducing the H$_2$O activity \dots \\
\hline
\end{tabularx}
\caption{
A case study of \textsc{REPAIR} generating fact-verified triplets ($q_\text{new}$, $d^+$, $d^-$) to resolve \myhl{long-tail concept} confusion in the materials science and biomedical domains.}
\label{tab:data_expansion_main}
\end{table*}

Table \ref{tab:maintable2} highlights the robust generalization of \textsc{REPAIR} across specialized  material and biomedical downstream tasks. With mid-sized parameters, \textsc{BMRetriever-2B}~\cite{xu2024bmretriever} exhibits slightly higher overall performance in the biomedical domains, these marginal gaps are primarily attributable to its larger scale and an exhaustive multi-task instruction fine-tuning. That is, \textsc{BMRetriever-2B} is explicitly aligned with downstream tasks by aggregating  human-annotated datasets and synthesizing task-specific scenarios to adapt to various input formats.

In contrast, \textsc{REPAIR} achieves exceptional generalization without this task-specific engineering. Not only does \textsc{REPAIR-1.5B} directly outperform the larger \textsc{BMRetriever-2B} on several specific tasks, but our 500M and 7B variants  consistently achieves the best performance across all evaluated tasks. By simply utilizing a unified query format, \textsc{REPAIR} eliminates the overhead of curating diverse query-passage pairs. \textsc{REPAIR} can seamlessly adapts to diverse and complex scenarios, including question answering to entity linking, demonstrating remarkable parameter efficiency. Our fact-verified grounding approach can establish a universally adaptable semantic space, rather than memorizing downstream task instructions.

%여기선 우리의 단점을 최대한 가리ㅈ거나 정다오하하고 결론은 이 걸 통해서 윌의 방법론의 강정ㅁ을 정당화하거나 강점을 보여주도록

\subsection{Ablation Studies and Analyses}
\label{sec:ablation}
We perform ablation studies to isolate the contributions of two key design choices in \textsc{REPAIR}, iterative refinement and fact-verified data augmentation. We also conduct an empirical analysis to validate the single-positive assumption underlying our diagnosis stage. Detailed quantitative results are provided in Appendix~\ref{app:Ablation}.

%We perform ablation studies and hyperparameter analyses to isolate the contribution of key design choices in \textsc{REPAIR}. All experiments are conducted under identical training and evaluation settings, differing only in the component under investigation. All detailed results are provided in Appendix~\ref{app:Ablation}.
%Appendix~\ref{app:DetaileRefinement}.
%Appendix~\ref{appendix:query_selection_ratio}.
% Detailed quantitative results are provided in Appendix~\ref{sec:appendix_k}.
\paragraph{Effect of Iterative Refinement.}
\label{ablation:iteration}
Figure~\ref{fig:abb_iter} evaluates iterative self-diagnosis across \textsc{REPAIR} models of different sizes (500M, 1.5B, 7B). By recomputing the margin landscape ${\Delta_\theta(q)}$ at each step, our approach dynamically resolves long-tail failure modes, yielding consistent performance improvements across all model capacities as iterations progress. Two iterations raise the average nDCG@10 by $10.0\%$ to $11.4\%$ over the seed-only baseline, as annotated above the bars. While performance continues to rise with additional iterations, the marginal gains progressively diminish, whereas the per-iteration cost stays constant (Appendix~\ref{app:cost}). Given that the most substantial improvements occur within the first two rounds, we set the default number of iterations to two for all of our experiments.

\paragraph{Robustness of the Selection Parameters.}
To avoid tuning the pipeline for each model, we use one setting for every
model size and every iteration: $p=40\%$ and $k=30$. Both values come from
separate measurements. Raising $p$ beyond $40\%$ finds few new concepts
(Tables~\ref{tab:concept_extraction_ratio} and~\ref{tab:detail_qyerq_margin_score}),
and six measures of negative quality all point to $k=30$
(Figure~\ref{fig:appendix_k_fine}). With this one setting, the average
nDCG@10 improves at every iteration for all three model sizes
($0.530 \to 0.547 \to 0.583$ for 500M, $0.546 \to 0.586 \to 0.607$ for 1.5B,
and $0.559 \to 0.589 \to 0.623$ for 7B), and it keeps improving up to the
fourth iteration (Table~\ref{tab:abblationiter}). One setting is therefore
enough across model sizes and iterations, with no re-tuning. % leaving the exploration of cost-effective, heavily scaled iterative refinement as future work. 

    \paragraph{Effect of Fact-Verified Data Beyond Model Capacity.} To  verify that \textsc{REPAIR}'s improvements stem from our data refinement rather than the Qwen2.5's 	inherent capacity, we isolate the effect of the augmented training data. As shown in Figure \ref{fig:qwen}, we train the Qwen2.5 models entirely on the augmented dataset used in a strong baseline, \textsc{BMRetriever}. Across all parameter scales, these models yield lower retrieval performance compared to \textsc{REPAIR}. This confirms that our core approach, resolving long-tail concept confusion through API-guided, fact-verified iterative refinement, is the fundamental driver of  enhanced scientific retrieval, proving that the quality of well-curated data outweighs the backbone capacity. We reach the same conclusion when we replace the backbone instead of the data, applying our pipeline to four backbones from different model families (Appendix~\ref{app:cross_backbone}).

\paragraph{Validity of the Single-Positive Assumption in Diagnosis.}
\begin{comment}
    To efficiently isolate long-tail confusions, we treat top-$k$ retrieved documents as negatives against a single positive. Following prior work \cite{cohan2020specter}, we verified this via direct citation relationships, a rigorous proxy for semantic equivalence. Across our 4M augmented dataset for the 500M, 1.5B, and 7B models, citation overlap was $<0.0001\%$. Because scientific texts are highly fact-sensitive, these lexically similar, uncited documents are overwhelmingly true hard negatives, virtually eliminating false negatives.
\end{comment}

To efficiently isolate long-tail confusions, our diagnosis stage extracts distractor concepts by treating the retrieved documents as negatives against a single positive. To ensure false negatives do not skew this diagnosis, we analyze the direct citation relationships between the anchor positives and the retrieved negatives. In scientific literature, direct citation relationships are established as a rigorous proxy for true semantic equivalence \cite{cohan2020specter}. Our analysis reveals an overwhelmingly low citation overlap of just $<0.00001\%$. For comparison, we ran the same check on SciDocs pairs that are known to cite each other, and only $5.80\%$ of them showed a citation link. This is the highest rate our lookup can detect, and our hard negatives fall far below it, at the same level as randomly paired documents (Appendix~\ref{app:citation}). Since scientific text exhibits extreme fact-sensitivity, lexically similar documents without citation links are overwhelmingly true hard negatives rather than false negatives. Furthermore, as our concept mining statistically aggregates signals across a large query set, this infinitesimally small noise is heavily diluted. This confirms that our approach robustly captures genuine diagnostic signals without contamination.

Finally, beyond the nine benchmarks of Tables~\ref{tab:mainresult} and~\ref{tab:maintable2}, \textsc{REPAIR} retains its advantage on three held-out scientific benchmarks spanning multi-aspect scientific IR, physics community QA, and broad-coverage science QA (Appendix~\ref{app:extra_bench}).

\subsection{Case study}
\label{sec:Casestu}

Table \ref{tab:data_expansion_main} demonstrates how \textsc{REPAIR} resolves the confusion of long-tail concepts using augmented fact-verified triplets ($q_\text{new}$, $d^+$, $d^-$). By grounding identified concepts ($\mathcal{C}_\text{conf}$), the framework generates targeted queries that probe precise yet underrepresented distinctions. For example, grounding \emph{Pb-based perovskite} constructs a query that pairs a positive document ($d^+$) detailing \emph{dynamic symmetry breaking} with a fact-contrastive hard negative ($d^-$) addressing \emph{static Rashba splitting}. Similarly, grounding \emph{poly(dl-lactic acid)} yields a query that retrieves a positive document ($d^+$) detailing its \emph{size-dependent hydrolytic degradation}, while isolating a fact-contrastive negative ($d^-$) that discusses its chemical formula \emph{C$_3$H$_6$O$_3$} in the unrelated context of \emph{zinc-ion batteries (AZIBs)}. This concept-driven, evidence-based expansion enables the retriever to resolve fine-grained factual distinctions.

\section{Conclusion}
We  presented \textsc{REPAIR}, a self-evolving framework that has effectively addressed the persistent challenges of long-tailed entities and high fact-sensitivity in scientific retrieval. 
By grounding iterative refinement in API-guided evidence, we  demonstrated that diagnosing specific knowledge gaps outperforms indiscriminate data augmentation.
While we focused on materials science and biomedicine, moving \textsc{REPAIR}
to a new domain is straightforward. Stages I and III depend only on the
retriever and its training data, so they transfer unchanged, and only the
evidence source in Stage II has to be replaced. Within science this means
plugging in resources such as ChEMBL, the NIST WebBook, or NASA ADS. Beyond
it, the same recipe applies to any field that has an authoritative database,
such as USPTO for patents or SEC EDGAR for finance. We leave a full study of
physics, engineering, and the social sciences to future work.

Ultimately, our work established a new paradigm, proving that integrating external verification into the training loop is essential for trustworthy knowledge discovery, and encouraging future research to prioritize rigorous factual verification.

\section*{Limitations}
\textsc{REPAIR} improves retrieval through an iterative loop, and each
iteration carries an additional cost. In practice this cost is bounded,
since the loop saturates at the second iteration across all model scales;
we report the per-stage breakdown in Appendix~\ref{app:cost}. The other
side of that saturation is a limitation: deeper iterations buy little, with
average nDCG@10 improving by at most $+0.005$ beyond the second iteration.
Simply extending the loop is therefore not a route to further gains, and
widening the evidence expansion within each iteration is a more promising
direction we leave to future work.

A second limitation is that \textsc{REPAIR} is bounded by its external
verifiers. Concepts the scientific APIs cannot resolve are discarded rather
than approximated (\S\ref{subsec:expansion}), which keeps supervision
factual but leaves those regions of the long tail untouched. Coverage thus
extends only as far as the available scientific resources do, and reaching
domains beyond materials science and biomedicine requires plugging in an
appropriate API for that field.

\section*{Acknowledgements}
%사람중심, 경량화, 대학중점연구소, AI펠로우십, AI대학원
This work was supported by Institute of Information \& communications Technology Planning \& Evaluation (IITP) grant funded by the Korea government (MSIT) (No.~RS-2022-II220156, Fundamental research on continual meta-learning for quality enhancement of casual videos and their 3D metaverse transformation), 
Institute of Information \& communications Technology Planning \& Evaluation (IITP) grant funded by the Korea government(MSIT) (No.RS-2026-25524173, Ultro-Long-Term Hierarchical Memory and Reasoning Architecture for Next-Generation Omnimodal Agents),
Basic Science Research Program through the National Research Foundation of Korea(NRF) funded by the Ministry of Education(RS-2023-00274280),
the Institute of Information \& Communications Technology Planning \& Evaluation(IITP) grant funded by the Korea government(MSIT) (RS-2025-25442338, AI star Fellowship Support Program(Seoul National Univ.)),
Institute of Information \& communications Technology Planning \& Evaluation (IITP) grant funded by the Korea government (MSIT) (No.~RS-2021-II211343, Artificial Intelligence Graduate School Program (Seoul National University))
, and
the AI Seoul Tech Research Support Program of the Seoul
Future Foundation.
Gunhee Kim is the corresponding author.

\begin{comment}
    Furthermore, our framework leverages external knowledge modules, specifically entity extractors (e.g., \textsc{MATDETECTOR}) and structured databases (e.g., PubChem). While these resources are empirically validated standards that ensure reliable diagnosis, they inherently bound performance by their coverage and schema alignment—particularly regarding emerging concepts or rare terminology. Crucially, however, the core mechanism of \emph{Self-Diagnosed Factual Expansion} is designed to be model- and domain-agnostic. The reliance on specific tools represents a modular implementation rather than a functional constraint, indicating that \textsc{REPAIR} can be readily extended to other scientific domains by integrating generic en아니 구tity linkers or alternative domain-specific APIs.

\end{comment}

\begin{comment}
    api 나 DB 다양핫게

\end{comment}

\begin{comment}
    This validates the necessity of the closed-loop design rather than a one-shot expansion strategy. 
여기에서 REPAIR-500M의 성능을 봤을떄 iter가 올라갈수록 선능이 계속 올라가긴하지만 초반에비해 완만하게 증가 (그래도 올라간다는걸 강조해서) 그래서 컴퓨테이셔널 코스트 등을 고려했을떄  iter 2 로 설정함. 하드웨어의 한계가 있었지만, 퓨처웍으로 데이터를 늘리는것도 좋을거임
\end{comment}


%\section*{Acknowledgments}

\bibliography{custom}
\clearpage

\appendix

\begin{table*}[!t]
\centering
\arraybackslash
\fontsize{8}{10.2}\selectfont % 9pt 글씨, 11pt 줄 간격
\setlength{\tabcolsep}{3pt} % 기본값은 6pt
\renewcommand{\arraystretch}{1.5} % 행 높이를 1.5배로 늘림
\resizebox{\textwidth}{!}{%
{\footnotesize
\begin{tabular}{c|c|c|l}
\hline
Domain & Dataset & Size & Line \\
\hline
\multirow{3}{*}{Material} & Mat2Vec \cite{tshitoyan2019unsupervised} & 1.5 M & https://github.com/materialsintelligence/mat2vec/ \\
 & MatSciBERT \cite{gupta2022matscibert} & 0.1 M & https://github.com/M3RG-IITD/MatSciBERT \\
 & MatBERT \cite{trewartha2022quantifying} & 2 M & https://github.com/lbnlp/MatBERT \\
 \hline 
\multirow{4}{*}{BioMedical} & S2ORC \cite{lo2020s2orc} & 600K & https://github.com/allenai/s2orc \\
 & Meadow \cite{wang-etal-2020-cord} & 460k & https://huggingface.co/datasets/medalpaca/medical\_meadow\_cord19 \\
 & Textbooks \cite{xiong2024benchmarking} & 50K & https://huggingface.co/datasets/MedRAG/textbooks \\
 & LitCovid \cite{chen2021litcovid} & 70K & https://huggingface.co/datasets/KushT/LitCovid\_BioCreative
\\ \hline
\end{tabular}
}}
\caption{Statistics of the public scientific corpora used for model initialization, categorized by domain.}
\label{tab:seed_data}
\end{table*}

\section{Details of Implementation and Setup}
\label{sec:implementation}

\subsection{Initial Corpus Construction}
\label{Initial Corpus Construction}
We prioritize data quality and domain breadth over sheer scale. Unlike standard baselines that rely on massive, noisy web-crawled corpora, we constructed a compact yet highly diverse dataset spanning a wider range of scientific disciplines, specifically integrating large-scale biomedical benchmarks with our newly constructed materials science data (see Table~\ref{tab:seed_data}). Although smaller in total volume compared to general-domain pre-training corpora, this curated mixture undergoes rigorous cleaning to ensure superior density of scientific information.

For the materials science domain, which specifically lacks unified public resources, we crawled documents via DOIs and addressed the substantial inconsistency in notation (e.g., \textit{$\alpha$-Fe$_2$O$_3$} vs. \textit{alpha-Fe$_2$O$_3$}). We applied a materials-aware normalization pipeline adapted from the MatSciBERT framework, including NFKC normalization, HTML entity mapping, and chemical formula hyphen reconnection. Crucially, we deliberately excluded standard normalization steps that would destroy materials-specific semantics, such as replacing numbers with placeholders or normalizing stoichiometric formulas (e.g., \textit{Ni$_{0.5}$Fe$_{0.5}$} $\to$ \textit{FeNi}).

Finally, we maximized data efficiency through strict quality filtering and consistent instruction formatting. We removed entries with missing metadata, as well as those exceeding context limits or lacking sufficient information. To leverage the instruction-following capabilities of the base model, we format every query $q$ with a specific task instruction:
\begin{center}
\texttt{``Given a query, retrieve passages that are relevant to the query.\textbackslash nQuery: \{text\} \{eos\}''}
\end{center}
This results in a refined corpus that is surface-consistent yet semantically precise, enabling the model to learn robust scientific representations from a smaller but more potent dataset.

\subsection{Details of Implementation}
\label{app:implementation}
All models are trained using PyTorch with Distributed Data Parallel (DDP) on two NVIDIA H200 GPUs. We adopt Qwen2.5-0.5B, Qwen2.5-1.5B, and Qwen2.5-7B \cite{qwen2} as backbone encoders, initialized from publicly released checkpoints. Training is performed in \texttt{bfloat16} precision with gradient checkpointing enabled to reduce memory consumption. We apply parameter-efficient fine-tuning with LoRA \cite{DBLP:conf/iclr/HuSWALWWC22}, using rank $r=16$, scaling factor $\alpha=32$, and dropout rate 0.05, and update only the LoRA parameters while keeping the backbone frozen. Optimization is carried out using AdamW \cite{DBLP:conf/iclr/LoshchilovH19}, with a learning rate of $2\times10^{-5}$ for the 7B model, and training proceeds for two epochs with a global batch size of 256 across GPUs. Input queries and passages are tokenized with a maximum sequence length of 512 and encoded using an EOS-based last-token pooling strategy to obtain fixed-dimensional representations. The retriever is trained with an InfoNCE contrastive objective, leveraging in-batch negatives as well as cross-device negatives enabled by DDP synchronization. Model checkpoints are saved periodically during training, and all hyperparameters are fixed across runs unless otherwise specified.

\subsubsection{Verified Query Generation}

To generate challenging queries in \emph{Expansion} (\S\ref{subsec:expansion}), from the detected long-tail scientific concepts, we employ a prompt-based query generation strategy. Given a target concept identified during self-diagnosis, we instruct a large language model to produce a single, specific research-oriented query grounded in materials science. The prompt template used for query generation is shown below.

\begin{promptbox}
Your task is to generate a single research query that is inherently ambiguous or difficult to resolve for standard retrieval models. The query should sound like a real paper title or research question — using indirect, context-dependent, or metaphorical language — so that a retrieval model cannot easily find the answer without deep understanding. \\

\textbf{Concept 1:} \{Concept\_1\} (Attributes: \{Attributes\_1\})\\
\textbf{Concept 2:} \{Concept\_2\} (Attributes: \{Attributes\_2\})\\ 
\textbf{Concept 3:} \{Concept\_3\} (Attributes: \{Attributes\_3\})  \\ \\
\textbf{Query:}

\end{promptbox}

% =========================================================
\subsection{Statistical Characterization of the Scientific Long Tail}
\label{app:longtail_stats}

Figure~\ref{fig:intro_overview}(b) contrasts a scientific corpus with MS
MARCO. To check that this contrast does not depend on a single reference
corpus, we compare five frequency distributions: scientific concepts, the
science corpus as a whole, the general words inside that corpus, and two
independent general-domain corpora, MS MARCO~\cite{bajaj2016ms} and
WikiText-103~\cite{merity2016pointer}. The same regular-expression tokenizer
is applied to all five so that the counts are comparable.

Table~\ref{tab:zipf} reports the rank-frequency statistics. The tail of the
scientific concepts is more than twice as flat as either general-domain
corpus, with a log-log Zipf slope of $0.82$ against $1.82$ for MS MARCO and
$1.76$ for WikiText-103. The gap is even clearer in how rare the terms are:
$62.4\%$ of scientific concepts appear exactly once and $93.9\%$ appear five
times or fewer in a 36.2M-token corpus, against $37$--$40\%$ and $67$--$69\%$
for the two general-domain corpora. In other words, there is almost no dense
head from which a retriever could learn these concepts, which is why
resampling the training data internally cannot fix the problem and why the
Expansion stage draws evidence from outside the corpus.

Table~\ref{tab:ks} tests the separation directly. Against both general-domain
corpora the two-sample Kolmogorov-Smirnov distance is $0.29$--$0.31$ with a
p-value below $10^{-300}$. The control comparison, scientific concepts against
the science corpus they are drawn from, is far smaller at $D=0.06$. The
separation is therefore between the scientific and general domains, not
between two samples of the same corpus.

\begin{table}[!t]
\centering
\fontsize{6.5}{8}\selectfont
\setlength{\tabcolsep}{3pt}
\renewcommand{\arraystretch}{1.2}
\resizebox{\columnwidth}{!}{%
\begin{tabular}{l|cc|ccc}
\hline
\multirow{2}{*}{Distribution} & Zipf slope & Zipf $\alpha$ & Hapax & freq $\le5$ & freq $\le10$ \\
 & $s$ (R$^2$) & (MLE) & (\%) & (\%) & (\%) \\ \hline
Science Concepts & 0.823 (0.904) & 1.862 & 62.43 & 93.87 & 96.52 \\
Science Corpus (overall) & 1.097 (0.922) & 1.751 & 56.51 & 89.72 & 93.49 \\
General Words (in-science) & 1.446 (0.962) & 1.606 & 45.15 & 82.12 & 87.90 \\ \hline
MS MARCO & 1.818 (0.981) & 1.455 & 37.01 & 66.79 & 75.13 \\
WikiText-103 & 1.761 (0.983) & 1.476 & 39.55 & 68.62 & 77.26 \\ \hline
\end{tabular}}
\caption{Rank-frequency statistics of five distributions under an identical
tokenizer. A smaller Zipf slope $s$ means a heavier tail, and Hapax is the
share of types that occur exactly once.}
\label{tab:zipf}
\end{table}

\begin{table}[!t]
\centering
\fontsize{7}{8.5}\selectfont
\setlength{\tabcolsep}{4pt}
\renewcommand{\arraystretch}{1.2}
\resizebox{\columnwidth}{!}{%
\begin{tabular}{l|cccc}
\hline
Reference & KS $D$ & KS $p$ & Wasserstein-1 & JS div. \\ \hline
MS MARCO & 0.3127 & $<10^{-300}$ & 0.4513 & 0.0751 \\
WikiText-103 & 0.2936 & $<10^{-300}$ & 0.4090 & 0.0668 \\ \hline
Science Corpus (control) & 0.0592 & $<10^{-300}$ & 0.0745 & 0.0039 \\ \hline
\end{tabular}}
\caption{Distributional distance from Science Concepts, measured in
$\log_{10}$ frequency space. The last row compares the scientific concepts
with the corpus they are drawn from and serves as a within-domain control.}
\label{tab:ks}
\end{table}
% =========================================================

% =========================================================

\section{Details of Evaluation}

\subsection{Baselines for Retrieval Tasks}
\label{app:baselines}
In this section, we provide detailed descriptions of the baseline models used in our experiments. A comprehensive summary of their architectural characteristics and methodological components, alongside our proposed REPAIR framework, is provided in Table \ref{tab:baseline_model}.

\paragraph{Sparse Retrieval Models.} 
Sparse retrieval approaches estimate relevance by matching keywords between queries and documents.
\begin{itemize}
    \item \textbf{BM25}~\cite{robertson2009probabilistic} serves as the standard probabilistic baseline for lexical retrieval. It utilizes a term-frequency inverse-document-frequency (TF-IDF) based scoring function to compute similarity scores between high-dimensional sparse vectors, effectively weighting term importance.
\end{itemize}

\paragraph{Dense Retrieval Models.} 
Dense retrieval models encode queries and documents into continuous vector spaces to capture semantic relationships. We evaluate models across three distinct scales:

\begin{itemize}
    \item \textbf{Contriever}~\cite{izacardunsupervised} is a dual-encoder model (110M) trained via unsupervised contrastive learning. It leverages a massive corpus comprising data from Wikipedia and CC-Net to learn robust representations without labeled supervision.
    \item \textbf{Dragon}~\cite{DBLP:conf/emnlp/LinALOLMY023} is a BERT-base scale model (110M) that adopts a progressive training strategy. It utilizes diverse supervision signals and data augmentation techniques to enhance general retrieval capabilities.
    \item \textbf{SPECTER 2.0}~\cite{singh2023scirepeval} is specifically tailored for scientific document representation (110M). It employs a multi-task training objective that covers various scientific tasks, allowing the model to generate embeddings adaptable to different formats and downstream applications.
    \item \textbf{SciMult}~\cite{zhang2023pre} is a domain-specialized retriever (110M) for scientific literature. It integrates instruction tuning within a multi-task contrastive learning framework to better align representations with scientific query intents.
    \item \textbf{MedCPT}~\cite{jin2023medcpt} focuses on biomedical information retrieval (220M). Its representations are learned from a large-scale dataset of 255 million user search logs from PubMed, effectively capturing the semantics of medical queries and documents.
    \item \textbf{InstructOR-L}~\cite{su2023one} is an instruction-finetuned model (335M) capable of generating task-specific embeddings. By conditioning on natural language instructions, it adapts to diverse domains without further fine-tuning.
    \item \textbf{E5-Large-v2}~\cite{wang2022text} employs a two-stage training pipeline (335M): initial contrastive pre-training on weakly labeled text pairs followed by supervised fine-tuning on high-quality datasets with mined hard negatives.
    \item \textbf{BGE-Large}~\cite{chen2024m3} is a strong baseline (335M) trained with a multi-stage approach similar to E5 but enhanced by improved negative sampling and a diverse training mixture.
    \item \textbf{\textsc{BMRetriever}-410M}~\cite{xu2024bmretriever} is the compact variant of a retrieval family tailored for biology and medicine. It is pre-trained on extensive domain-specific corpora and subsequently fine-tuned using augmented data synthesized by Large Language Models.
    \item \textbf{DRAMA-L}~\cite{ma2025drama} represents a lightweight baseline designed for efficient retrieval, balancing performance with computational constraints.
\end{itemize}

\begin{table*}[!t]
\centering
\arraybackslash
\fontsize{6}{6}\selectfont % 9pt 글씨, 11pt 줄 간격
\setlength{\tabcolsep}{3pt} % 기본값은 6pt
\renewcommand{\arraystretch}{2} % 행 높이를 1.5배로 늘림
\resizebox{\textwidth}{!}{%

\renewcommand{\arraystretch}{1.5}
\begin{tabular}{l|ccc|cccccc}
\hline
Method & Backbone & Scale & Domain & \begin{tabular}[c]{@{}c@{}}Contra\\      Pretrain.\end{tabular} & \begin{tabular}[c]{@{}c@{}}Data\\      Aug.\end{tabular} & \begin{tabular}[c]{@{}c@{}}Hard\\      Neg.\end{tabular} & \begin{tabular}[c]{@{}c@{}}Iter\\      Refine.\end{tabular} & \begin{tabular}[c]{@{}c@{}}Confus\\      Diag.\end{tabular} & \begin{tabular}[c]{@{}c@{}}Fact\\      Verif.\end{tabular} \\ \hline
BM25 & - & - & General & \xmark & \xmark & \xmark & \xmark & \xmark & \xmark \\
Contriever (\citeyear{izacardunsupervised}) & BERT-base (\citeyear{ni2022large}) & 110M & General & \textcolor{ForestGreen}{\cmark} & \textcolor{ForestGreen}{\cmark} & \textcolor{ForestGreen}{\cmark} & \xmark & \xmark & \xmark \\
Dragon (\citeyear{DBLP:conf/emnlp/LinALOLMY023}) & BERT-base (\citeyear{ni2022large}) & 110M & General & \xmark & \textcolor{ForestGreen}{\cmark} & \textcolor{ForestGreen}{\cmark} & \xmark & \xmark & \xmark \\
SPECTER 2.0 (\citeyear{singh2023scirepeval}) & SciBERT (\citeyear{beltagy2019scibert}) & 110M & Scientific & \textcolor{ForestGreen}{\cmark} & \xmark & \textcolor{ForestGreen}{\cmark} & \xmark & \xmark & \xmark \\
SciMult (\citeyear{zhang2023pre}) & PubMedBERT (\citeyear{gu2021domain}) & 110M & Scientific & \textcolor{ForestGreen}{\cmark} & \xmark & \textcolor{ForestGreen}{\cmark} & \xmark & \xmark & \xmark \\
MedCPT (\citeyear{jin2023medcpt}) & PubMedBERT (\citeyear{gu2021domain}) & 220M & Biomedical & \xmark & \textcolor{ForestGreen}{\cmark} & \textcolor{ForestGreen}{\cmark} & \xmark & \xmark & \xmark \\
InstructOR-L (\citeyear{su2023one}) & GTR-Large (\citeyear{ni2022large}) & 335M & General & \xmark & \textcolor{ForestGreen}{\cmark} & \textcolor{ForestGreen}{\cmark} & \xmark & \xmark & \xmark \\
E5-Large-v2† (\citeyear{wang2022text}) & BERT-large (\citeyear{ni2022large}) & 660M & General & \textcolor{ForestGreen}{\cmark} & \textcolor{ForestGreen}{\cmark} & \textcolor{ForestGreen}{\cmark} & \xmark & \xmark & \xmark \\
BGE-Large‡ (\citeyear{chen2024m3}) & RoBERTa-large (\citeyear{conneau2020unsupervised}) & 895M & General & \textcolor{ForestGreen}{\cmark} & \textcolor{ForestGreen}{\cmark} & \textcolor{ForestGreen}{\cmark} & \xmark & \xmark & \xmark \\
\textsc{BMRetriever}-410M (\citeyear{xu2024bmretriever}) & Pythia-410M (\citeyear{biderman2023pythia}) & 410M & Biomedical & \textcolor{ForestGreen}{\cmark} & \textcolor{ForestGreen}{\cmark} & \textcolor{ForestGreen}{\cmark} & \xmark & \xmark & \xmark \\
InstructOR-XL (\citeyear{su2023one}) & GTR-XL (\citeyear{ni2022large}) & 1.5B & General & \xmark & \textcolor{ForestGreen}{\cmark} & \textcolor{ForestGreen}{\cmark} & \xmark & \xmark & \xmark \\
GTR-XL (\citeyear{ni2022large}) & T5-XL    (\citeyear{vaswani2017attention}) & 1.2B & General & \textcolor{ForestGreen}{\cmark} & \textcolor{ForestGreen}{\cmark} & \textcolor{ForestGreen}{\cmark} & \xmark & \xmark & \xmark \\
GTR-XXL (\citeyear{ni2022large}) & T5-XXL (\citeyear{vaswani2017attention}) & 4.8B & General & \textcolor{ForestGreen}{\cmark} & \textcolor{ForestGreen}{\cmark} & \textcolor{ForestGreen}{\cmark} & \xmark & \xmark & \xmark \\
SGPT-1.3B (\citeyear{muennighoff2022sgpt}) & GPT-Neo (\citeyear{andonian2023gpt}) & 1.3B & General & unk & \textcolor{ForestGreen}{\cmark} & \xmark & \xmark & \xmark & \xmark \\
SGPT-2.7B (\citeyear{muennighoff2022sgpt}) & GPT-Neo (\citeyear{andonian2023gpt}) & 2.7B & General & unk & \textcolor{ForestGreen}{\cmark} & \xmark & \xmark & \xmark & \xmark \\
\textsc{BMRetriever}-2B (\citeyear{xu2024bmretriever}) & Gemma (\citeyear{team2024gemma}) & 2B & Biomedical & \textcolor{ForestGreen}{\cmark} & \textcolor{ForestGreen}{\cmark} & \textcolor{ForestGreen}{\cmark} & \xmark & \xmark & \xmark \\
DRAMA-1B (\citeyear{ma2025drama}) & LLaMA-3.2-1B & 1B & General & \xmark & \textcolor{ForestGreen}{\cmark} & \textcolor{ForestGreen}{\cmark} & \xmark & \xmark & \xmark \\
Llama2Vec (\citeyear{li2024llama2vec}) & LLaMA-2-7B & 7B & General & \textcolor{ForestGreen}{\cmark} & \textcolor{ForestGreen}{\cmark} & \textcolor{ForestGreen}{\cmark} & \xmark & \xmark & \xmark \\
RepLLaMA (\citeyear{ma2024fine}) & LLaMA-2-7B & 7B & General & \xmark & \textcolor{ForestGreen}{\cmark} & \textcolor{ForestGreen}{\cmark} & \xmark & \xmark & \xmark \\
LLM2Vec (\citeyear{behnamghader2024llm2vec}) & Mistral-7B & 7B & General & \textcolor{ForestGreen}{\cmark} & \textcolor{ForestGreen}{\cmark} & \textcolor{ForestGreen}{\cmark} & \xmark & \xmark & \xmark \\
E5-Mistral (\citeyear{wang2024improving}) & Mistral-7B & 7B & General & \xmark & \textcolor{ForestGreen}{\cmark} & \textcolor{ForestGreen}{\cmark} & \xmark & \xmark & \xmark \\
CPT-text-XL (\citeyear{neelakantan2022text}) & GPT (\citeyear{brown2020language}) & 175B & General & unk & unk & \xmark & \xmark & \xmark & \xmark \\
\textsc{BMRetriever}-7B (\citeyear{xu2024bmretriever}) & BioMistral (\citeyear{labrak2024biomistral}) & 7B & Biomedical & \textcolor{ForestGreen}{\cmark} & \textcolor{ForestGreen}{\cmark} & \textcolor{ForestGreen}{\cmark} & \xmark & \xmark & \xmark \\
Promptriever (\citeyear{DBLP:conf/iclr/WellerDLPZH25}) & llama2-7b (\citeyear{touvron2023llama}) & 7B & General & \textcolor{ForestGreen}{\cmark} & \textcolor{ForestGreen}{\cmark} & \textcolor{ForestGreen}{\cmark} & \xmark & \xmark & \xmark \\ \hline
REPAIR-500M (ours) & Qwen2.5-0.5B (\citeyear{qwen2}) & 500M & Scientific & \textcolor{ForestGreen}{\cmark} & \textcolor{ForestGreen}{\cmark} & \textcolor{ForestGreen}{\cmark} & \textcolor{ForestGreen}{\cmark} & \textcolor{ForestGreen}{\cmark} & \textcolor{ForestGreen}{\cmark} \\
REPAIR-1.5B (ours) & Qwen2.5-1.5B (\citeyear{qwen2}) & 1.5B & Scientific & \textcolor{ForestGreen}{\cmark} & \textcolor{ForestGreen}{\cmark} & \textcolor{ForestGreen}{\cmark} & \textcolor{ForestGreen}{\cmark} & \textcolor{ForestGreen}{\cmark} & \textcolor{ForestGreen}{\cmark} \\
REPAIR-7B (ours) & Qwen2.5-7B (\citeyear{qwen2}) & 7B & Scientific & \textcolor{ForestGreen}{\cmark} & \textcolor{ForestGreen}{\cmark} & \textcolor{ForestGreen}{\cmark} & \textcolor{ForestGreen}{\cmark} & \textcolor{ForestGreen}{\cmark} & \textcolor{ForestGreen}{\cmark} \\ \hline
\end{tabular}}
\caption{Comprehensive comparison of baseline retrieval models and the proposed REPAIR framework. The table delineates backbone architectures, model scales, target domains, and specific training methodologies. Methodological components are abbreviated as follows: Contra Pretrain. (Contrastive Pretraining), Data Aug. (Data Augmentation), Hard Neg. (Hard Negative Mining), Iter Refine. (Iterative Refinement), Confus Diag. (Confusion Diagnosis), and Fact Verif. (Factual Verification).}
  \label{tab:baseline_model}
\end{table*}

\begin{itemize}
    \item \textbf{InstructOR-XL}~\cite{su2023one} scales the instruction-based training methodology to 1.5B parameters, offering improved generalization and instruction-following capabilities compared to its smaller counterpart.
    \item \textbf{GTR-XL / GTR-XXL}~\cite{ni2022large} are Generalizable T5-based Retrievers. Initialized from T5, they undergo pre-training on community QA pairs followed by fine-tuning on NQ and MS MARCO. We report results for the 1.2B and 4.8B variants.
    \item \textbf{SGPT-1.3B / SGPT-2.7B}~\cite{muennighoff2022sgpt} adapt decoder-only GPT architectures for symmetric search. By freezing the backbone and fine-tuning only the bias tensors and position-weighted pooling layers, they transform generative models into effective retrievers.
    \item \textbf{\textsc{BMRetriever}-2B}~\cite{xu2024bmretriever} scales the biomedical-focused architecture to 2 billion parameters, allowing for deeper semantic understanding of scientific texts.
    \item \textbf{DRAMA-1B}~\cite{ma2025drama} is the billion-scale iteration of the DRAMA series, providing a middle-ground baseline between efficiency and capacity.
\end{itemize}

\begin{itemize}
    \item \textbf{Llama2Vec}~\cite{li2024llama2vec} converts LLaMA-7B into a retriever using two novel pre-training tasks: Embedding-Based Auto-Encoding (EBAE) and Embedding-Based Next Sentence Prediction (EBNSP).
    \item \textbf{RepLLaMA}~\cite{ma2024fine} performs full fine-tuning of the LLaMA-7B model on MS MARCO, directly optimizing the generative backbone for passage retrieval tasks.
    \item \textbf{LLM2Vec}~\cite{behnamghader2024llm2vec} enables bidirectional attention in causal LLMs through masked next-token prediction. This unsupervised approach transforms standard LLMs into powerful text encoders.
    \item \textbf{E5-Mistral}~\cite{wang2024improving} initializes from Mistral-7B and is trained with a wide variety of synthetic data generated by LLMs, achieving state-of-the-art performance on the MTEB benchmark.
    \item \textbf{CPT-text-XL}~\cite{neelakantan2022text} is a web-scale contrastive model (175B). We include it as a reference point for performance achievable with massive-scale pre-training, rather than a direct comparison due to its size.
    \item \textbf{\textsc{BMRetriever}-7B}~\cite{xu2024bmretriever} is the largest model in its series, leveraging 7 billion parameters to maximize retrieval accuracy in specialized scientific domains.
    \item \textbf{Promptriever}~\cite{DBLP:conf/iclr/WellerDLPZH25} is a bi-encoder retrieval model initialized from an LLM backbone. Unlike standard retrievers, it is fine-tuned on a massive dataset of MS MARCO pairs augmented with instance-level instructions and ``instruction negatives, enabling it to follow complex, per-instance natural language prompts to dynamically adjust relevance criteria without further training.
\end{itemize}

\subsection{Evaluation Task and Dataset}
\label{app:taskdataset}
In this section, we provide detailed descriptions of the datasets employed in our experiments. We categorize these benchmarks into five primary retrieval-oriented groups: Information Retrieval (IR), Sentence Similarity, Question Answering (QA), Entity Linking, and Paper Recommendation.

\subsubsection{Information Retrieval}
Following prior work~\cite{xu2024bmretriever}, we evaluate passage retrieval performance in scientific and biomedical domains using four datasets from the BEIR benchmark~\cite{boteva2016full}. These benchmarks require retrieving relevant passages from corpora containing complex, terminology-intensive documents.

\paragraph{NFCorpus} \cite{boteva2016full}: A biomedical information retrieval dataset consisting of 323 natural-language queries related to nutrition facts, evaluated over a corpus of approximately 3.6K PubMed documents. The task is formulated as document retrieval, where models are given a question and are required to retrieve documents that best answer the query.

\paragraph{SciFact} \cite{wadden2020fact}: A scientific fact-verification dataset comprising 300 queries, where the task is to retrieve abstracts that provide supporting or refuting evidence for a given scientific claim. The corpus consists of approximately 5K scientific papers.

\paragraph{SciDocs} \cite{cohan2020specter}: A citation-oriented retrieval dataset consisting of 1K queries derived from scientific paper titles, evaluated over a corpus of 25K scientific articles. The task requires retrieving abstracts of papers that are cited by the given paper.

\paragraph{TREC-COVID} \cite{voorhees2021trec}: A biomedical information retrieval dataset focused on COVID-19-related literature, comprising 50 queries evaluated over a corpus of approximately 171K documents. Each query is associated with a dense set of relevant documents, averaging 493.5 per query, and the task requires retrieving documents that answer the given COVID-19 query.

We additionally evaluate on three scientific benchmarks that lie outside the
nine used in the main experiments, in order to probe generalization to
unseen task formats and to scientific subareas beyond materials science and
biomedicine (Appendix~\ref{app:extra_bench}). None of the three is used at
any point during training.

\paragraph{DORIS-MAE} \cite{wang2023doris}: A multidisciplinary scientific
document retrieval dataset built from computer science literature, in which
each of the 100 queries is a multi-sentence research summary decomposed into
several aspects. Relevance is graded over a corpus of 8,591 abstracts, and
the multi-aspect query format differs markedly from the single-intent
queries of the four benchmarks above.

\paragraph{CQA-physics} \cite{hoogeveen2015cqadupstack, thakur2021beir}: The
physics subforum of CQADupStack as distributed in BEIR, consisting of 1,039
community question-answering queries over a corpus of 38,316 posts. The task
requires retrieving duplicate or answer-bearing posts, and its informal,
user-written style contrasts with the formal scientific prose of the other
benchmarks.

\paragraph{SciQ} \cite{welbl2017crowdsourcing}: A broad-coverage science QA
dataset spanning physics, chemistry, biology, and earth science. We cast it
as retrieval by pairing each of the 884 test questions with the supporting
passage that contains its answer, over a corpus of 12,241 deduplicated
support passages.

\begin{table*}[!t]
\centering
\arraybackslash
\fontsize{3}{4}\selectfont % 9pt 글씨, 11pt 줄 간격
\setlength{\tabcolsep}{3pt} % 기본값은 6pt
\renewcommand{\arraystretch}{1.2} % 행 높이를 1.5배로 늘림
\resizebox{\textwidth}{!}{%
\begin{tabular}{l|cc|ccccc|c}
\hline
 &  &  & \multicolumn{4}{c}{Standard   IR} & Sent. Sim. &  \\ \cline{4-8}
\multirow{-2}{*}{Task} & \multirow{-2}{*}{Iter.} & \multirow{-2}{*}{\# PT Pairs} & NFCorpus & SciFact & SciDocs & COVID & BIOSSES & \multirow{-2}{*}{AVG.} \\ \hline
 & - & 2.5M & 0.321 & 0.623 & 0.173 & 0.751 & 0.781 & 0.530 \\
 & 1 & 3.5M & 0.343 & 0.651 & 0.187 & 0.749 & 0.803 & 0.547 \\
 & \cellcolor[HTML]{   EBF2F6}2 & \cellcolor[HTML]{   EBF2F6}4.0M & \cellcolor[HTML]{   EBF2F6}\textbf{0.376} & \cellcolor[HTML]{   EBF2F6}\textbf{0.680} & \cellcolor[HTML]{   EBF2F6}\textbf{0.196} & \cellcolor[HTML]{   EBF2F6}\textbf{0.812} & \cellcolor[HTML]{   EBF2F6}\textbf{0.853} & \cellcolor[HTML]{   EBF2F6}\textbf{0.583} \\

 & 3 & 4.5M  & 0.378 & 0.693 & 0.200 & 0.814 & 0.855 & 0.588 \\ 
\multirow{-5}{*}{REPAIR-500M} & 4 & 5.0M & 0.382 & 0.697 & 0.199 & 0.813 & 0.852 & 0.589  \\

\hline
 & - & 2.5M & 0.321 & 0.676 & 0.193 & 0.739 & 0.799 & 0.546 \\
 & 1 & 3.5M & 0.371 & 0.732 & 0.191 & 0.802 & 0.836 & 0.586 \\
 & \cellcolor[HTML]{   EBF2F6}2 & \cellcolor[HTML]{   EBF2F6}4M & \cellcolor[HTML]{   EBF2F6}\textbf{0.376} & \cellcolor[HTML]{   EBF2F6}\textbf{0.757} & \cellcolor[HTML]{   EBF2F6}\textbf{0.201} & \cellcolor[HTML]{   EBF2F6}\textbf{0.853} & \cellcolor[HTML]{   EBF2F6}\textbf{0.849} & \cellcolor[HTML]{   EBF2F6}\textbf{0.607} \\
 & 3 & 4.5M & 0.379 & 0.762 & 0.204 & 0.855 & 0.851 & 0.610 \\
\multirow{-5}{*}{REPAIR-1.5B} & 4 & 5.0M & 0.381 & 0.763 & 0.203 & 0.855 & 0.852 & 0.611 \\
 \hline

 & - & 2.5M & 0.370 & 0.683 & 0.210 & 0.724 & 0.808 & 0.559 \\
 & 1 & 3.5M & 0.355 & 0.772 & 0.222 & 0.772 & 0.827 & 0.589 \\
 & \cellcolor[HTML]{   EBF2F6}2 & \cellcolor[HTML]{   EBF2F6}4M & \cellcolor[HTML]{   EBF2F6}\textbf{0.413} & \cellcolor[HTML]{   EBF2F6}\textbf{0.789} & \cellcolor[HTML]{   EBF2F6}\textbf{0.227} & \cellcolor[HTML]{   EBF2F6}\textbf{0.842} & \cellcolor[HTML]{   EBF2F6}\textbf{0.846} & \cellcolor[HTML]{   EBF2F6}\textbf{0.623} \\
 & 3 & 4.5M & 0.416 & 0.793 & 0.229 & 0.844 & 0.847 & 0.626 \\
\multirow{-5}{*}{REPAIR-7B} & 4 & 5.0M & 0.417 & 0.794 & 0.229 & 0.844 & 0.848 & 0.627
 \\ \hline

\end{tabular}
}
    \caption{Comparison of retrieval performance across iterations and model scales. The highlighted row marks our default setting (Iter 2). Beyond it, the average nDCG@10 improves by at most $+0.005$ at any scale.}

  \label{tab:abblationiter}
\end{table*}

\subsubsection{Sentence Similarity.}
For sentence-level retrieval, we employ \textbf{BIOSSES} \cite{souganciouglu2017biosses}, a biomedical sentence similarity dataset consisting of 100 sentence pairs extracted from PubMed articles. Each pair is annotated by human experts with a similarity score on a 5-point scale, ranging from 0 (no semantic relation) to 4 (semantically equivalent). The task is formulated as sentence retrieval, where models are given a sentence and are required to retrieve sentences with the same meaning.

\subsubsection{Question Answering.}
We extend our evaluation to retrieval-augmented downstream tasks using three QA datasets:

\paragraph{iCliniq} \cite{DBLP:journals/corr/abs-2004-03329}: A biomedical conversational question answering dataset constructed from patient–clinician interactions collected from a public health forum, comprising approximately 7.3K questions and 7.3K responses. The task is formulated as retrieval-based QA, where models are given a question with conversational context and are required to retrieve responses that best answer the query.

\paragraph{ChemLit-QA} \cite{DBLP:journals/mlst/WellawatteGLBHBS25}: A literature-based scientific QA and Retrieval-Augmented Generation (RAG) benchmark. This dataset evaluates the model's ability to generate faithful and precise answers based on chemical literature contexts. It was rigorously validated by four experts with backgrounds in chemistry and chemical engineering. For our experiments, we specifically utilized the subsets categorized under biomedical and material science domains to align with our target tasks.

    \subsubsection{Entity Linking.}
To assess the model's capability in identifying and linking domain-specific concepts, we use \textbf{MeSH} \cite{lipscomb2000medical}, a biomedical entity linking benchmark designed to evaluate the identification and normalization of domain-specific concepts. The dataset comprises approximately 29.6K biomedical concepts and corresponding textual entries from the Medical Subject Headings (MeSH) thesaurus. The task is formulated as retrieval-based entity linking, where models are given a biomedical concept mention and are required to retrieve passages that define or correspond to the correct MeSH concept.

\subsubsection{Paper Recommendation.}
We evaluate retrieval performance on a paper recommendation task using the \textbf{RELISH} dataset~\cite{singh2023scirepeval, DBLP:journals/biodb/BrownCZ19}. The benchmark consists of approximately 3.2K query articles and a corpus of 191.2K PubMed abstracts. The task requires retrieving literature relevant to a given article, with relevance annotated using graded similarity scores ranging from 0 (not similar) to 2 (highly similar).

% =========================================================
\section{Details of Ablation Studies and Analyses}
\label{app:Ablation}
This section provides comprehensive experimental details and extended results for ablation studies introduced in \S\ref{sec:ablation}. Specifically, we further investigate the individual contributions of key design choices in \textsc{REPAIR} by presenting detailed analyses on the iterative refinement process (\S\ref{app:DetaileRefinement}), the low-margin query selection ratio (\S\ref{appendix:query_selection_ratio}), and the impact of the number of analyzed negatives $k$ (\S\ref{sec:appendix_k}).

% =========================================================

\subsection{Detailed Analysis of Iterative Refinement}
\label{app:DetaileRefinement}

Table \ref{tab:abblationiter} illustrates the performance trajectory across iterations. The primary driver of these gains is the resolution of long-tail concept confusion rather than inherent model capacity. To isolate this effect, we compare the same Qwen2.5 backbones trained on our refined data versus a strong baseline (BMRetriever). Across all parameter scales, models trained with \textsc{REPAIR} consistently outperform those trained on baseline datasets, proving that fact-verified data quality outweighs backbone size.

The efficacy of this refinement is further evidenced by the representational margin shift. For the 173 "persistent queries" that remained in the confusion set after Iteration 1, the average margin shifted from $-2.7 \times 10^{-3}$ to $+3.5 \times 10^{-3}$ in Iteration 2. This positive shift indicates that the iterative process successfully expands the model's embedding space to distinguish fine-grained scientific concepts that were previously collapsed. Consequently, the refinement process ensures that the model's improvements are grounded in factual differentiation rather than biased stagnation.

% =========================================================

\begin{table*}[!t]
\centering
\arraybackslash
\fontsize{5}{6}\selectfont % 9pt 글씨, 11pt 줄 간격
\setlength{\tabcolsep}{3pt} % 기본값은 6pt
\renewcommand{\arraystretch}{1.3} % 행 높이를 1.5배로 늘림
\resizebox{\textwidth}{!}{%
\begin{tabular}{cccccc}
\hline
{Selection Ratio ($p$)} & {Path A Unique} & {Path B Unique} & {Total Unique ($A \cup B$)} & {Marginal Increase ($\Delta$)} \\
\hline
5\%   & 122,041 & 29,387  & 145,252 & - \\
10\%  & 176,656 & 41,316  & 208,710 & +63,458 \\
15\%  & 219,701 & 50,758  & 258,541 & +49,831 \\
20\%  & 256,586 & 57,474  & 300,265 & +41,724 \\
25\%  & 288,057 & 64,563  & 336,695 & +36,430 \\
30\%  & {315,667} & {71,257}  & {368,971} & {+32,276} \\
35\%  & 338,714 & 76,586  & 396,400 & +27,429 \\
\rowcolor[HTML]{   EBF2F6} 40\% & \textbf{361,762} & \textbf{81,916}  & \textbf{422,183} & \textbf{+53,212} \\
45\%  & 379,969 & 88,431  & 444,437 & +22,254 \\
50\%  & 396,021 & 94,375  & 464,468 & +20,031 \\
55\%  & 410,067 & 99,867  & 482,036 & +17,568 \\
60\%  & 422,889 & 104,726 & 498,039 & +16,003 \\
65\%  & 435,795 & 109,335 & 514,462 & +16,423 \\
70\%  & 447,391 & 113,780 & 529,456 & +14,994 \\
75\%  & 457,645 & 115,514 & 541,093 & +11,637 \\
80\%  & 466,562 & 119,872 & 553,281 & +12,188 \\
85\%  & 474,922 & 124,829 & 565,367 & +12,086 \\
90\%  & 482,318 & 130,163 & 576,690 & +11,323 \\
95\%  & 488,931 & 136,145 & 587,547 & +10,857 \\
100\% & 495,452 & 146,199 & 601,254 & +13,707 \\
\hline
\end{tabular}
}
\caption{Number of unique long-tail concepts extracted across different query selection margins ($p$). The Total Unique ($A \cup B$) shows the footprint of epistemic uncertainty captured by the diagnosis stage. The marginal increase ($\Delta$) significantly drops after $p=40\%$, indicating diminishing returns. Selecting beyond this threshold primarily introduces well-resolved concepts that act as noise during the data expansion stage. }
\label{tab:concept_extraction_ratio}

\end{table*}

\begin{table*}[!t]
\centering
\footnotesize % 논문에서 허용되는 적절히 작은 폰트 사이즈
\setlength{\tabcolsep}{8pt} % 열 간격. 표가 너무 넓으면 4pt~6pt로 줄이세요.
\renewcommand{\arraystretch}{1.3} % 행 높이 1.6은 너무 넓을 수 있어 1.3으로 조정
\begin{tabular}{c|ccccc|c}
\hline
Selection Ratio ($p$) & NFCorpus & SciFact & SciDocs & COVID & BIOSSES & AVG. \\ \hline
10\% & 0.321 & 0.630 & 0.180 & 0.730 & 0.788 & 0.530 \\
20\% & 0.310 & 0.644 & 0.184 & 0.738 & 0.793 & 0.534 \\
30\% & 0.334 & 0.640 & 0.180 & 0.743 & 0.801 & 0.540 \\
\rowcolor[HTML]{   EBF2F6} 
40\% & \textbf{0.343} & \textbf{0.651} & \textbf{0.187} & \textbf{0.749} & \textbf{0.803} & \textbf{0.547} \\ 
50\% & 0.328 & 0.659 & 0.184 & 0.751 & 0.806 & 0.545 \\
60\% & 0.335 & 0.656 & 0.187 & 0.754 & 0.810 & 0.548 \\
70\% & 0.342 & 0.653 & 0.191 & 0.757 & 0.809 & 0.550 \\
80\% & 0.330 & 0.652 & 0.195 & 0.768 & 0.814 & 0.552 \\
90\% & 0.324 & 0.654 & 0.196 & 0.764 & 0.816 & 0.551 \\
100\% & 0.328 & 0.664 & 0.201 & 0.760 & 0.817 & 0.554 \\ \hline
\end{tabular}
\caption{Detailed retrieval performance (nDCG@10) across five target datasets at varying low-margin query selection ratios ($p$). The highlighted row ($p=40\%$) indicates the optimal threshold that provides a strong balance between performance and concept efficiency.}
\label{tab:detail_qyerq_margin_score}
\end{table*}

\begin{comment}
    Figure~\ref{fig:abb_Query} ablates the proportion $p$ of queries selected into the confusion set $\mathcal{Q}_{\mathrm{conf}}$ based on the retrieval margin $\Delta_\theta(q)$ for the 0.5B model at iteration 1, and analyzes its effect on both the scale of the diagnosed concept set and downstream retrieval performance.
As $p$ increases, the total number of diagnosed concepts $|C_t|$ grows monotonically, but with a highly non-linear rate across percentiles.
The concept set expands rapidly up to $p=40\%$, after which the increase becomes marginal despite continued inclusion of higher-margin queries.
Retrieval performance exhibits a similar saturation pattern, improving in the low-percentile regime but showing no substantial gains beyond $p=40\%$.
Accordingly, we fix $p=40\%$ in all experiments.
These results validate margin-based query selection as an effective mechanism for isolating the subset of training queries that are most informative for diagnosing and correcting long-tail retrieval errors. Detailed results are provided in Appendix~\ref{appendix:query_selection_ratio}.
\end{comment}

\subsection{Detail Analysis of Low-Margin Query Selection Ratio}
\label{appendix:query_selection_ratio}

To evaluate the effectiveness of our margin-based selection in concentrating diagnostic signals for long-tail errors, we tests the selection ratio $p$ based on $\Delta_\theta(q)$. In conjunction with this visual summary, Table~\ref{tab:concept_extraction_ratio} and Table~\ref{tab:detail_qyerq_margin_score} provide the complete empirical results supporting our choice to fix $p=40\%$.

\paragraph{Concept Extraction Scale and Diminishing Returns.}
Table~\ref{tab:concept_extraction_ratio} details the number of unique long-tail concepts extracted via Path A and Path B as the selection ratio $p$ increases from $5\%$ to $100\%$. The total number of unique concepts ($A \cup B$) demonstrates rapid initial growth. However, the marginal increase ($\Delta$) column clearly illustrates a point of diminishing returns. Up to $p=40\%$, the diagnosis stage efficiently extracts $422,183$ unique concepts. Beyond this threshold, increasing the ratio requires processing a significantly larger volume of queries, but the marginal discovery of novel concepts drops. This indicates that queries above the 40th percentile of the retrieval margin $\Delta_\theta(q)$ are largely well-resolved by the base retriever and contribute little to the footprint of epistemic uncertainty.

\paragraph{Downstream Retrieval Performance.}
Table~\ref{tab:detail_qyerq_margin_score} reports the exact nDCG@10 scores across the five individual target datasets (NFCorpus, SciFact, SciDocs, COVID, BIOSSES). The average nDCG@10 score rises steadily from $0.530$ at $p=10\%$ to $0.547$ at $p=40\%$. Beyond $p=40\%$, the performance exhibits a clear saturation effect. While processing $100\%$ of the queries yields the absolute maximum average of $0.554$, the gain from $p=40\%$ is minimal ($+0.007$). Given the substantial computational cost of the data expansion stage, introducing the remaining $60\%$ of queries primarily acts as noise. Therefore, $p=40\%$ provides an optimal balance, maximizing diagnostic value while maintaining high retrieval accuracy.

% =========================================================

% ============================================================
% Appendix: Detailed Analysis of the Number of Analyzed Negatives (k)
% Single-figure version (uses only v3f_fig3_fine_centered.png)
% ============================================================

\subsection{Detailed Analysis of the Number of Analyzed Negatives ($k$)}
\label{sec:appendix_k}

\paragraph{Setup and Motivation.} The core strength of the REPAIR framework lies in its ability to precisely isolate long-tail confusions without being polluted by semantic noise or irrelevant distractors. During the diagnosis stage, identifying the optimal number of analyzed top-ranked negatives ($k$) is critical: inspecting too few negatives might fail to capture systemic error patterns, while inspecting too many risks introducing semantic drift that degrades the factual fidelity of the extracted concepts.

To systematically justify the optimal boundary of $k=30$, we evaluate the neighborhood stability and negative hardness employing the 0.5B retriever at the initial iteration. For each confused query $q \in \mathcal{Q}_{\mathrm{conf}}$, we construct a ranked list of negatives $\mathcal{N}_k(q)=\{\hat d_1,\ldots,\hat d_k\}$ using the current encoder while strictly excluding the paired positive document $d^+$. By fixing the confusion selection ratio at $p=40\%$, we obtain confused queries and subsequently sweep the parameter $k$ across the set $\{5,10,15,\dots,100\}$. All measurements utilize cosine similarities between L2-normalized embeddings derived via end-of-sequence last-token pooling, which are efficiently computed through a cached top-100 retrieval matrix.

\subsubsection{Quantitative Analysis}
\label{sec:appendix_k_metrics}
The core objective of the REPAIR framework is to accurately diagnose the model's vulnerabilities by exposing it to genuine hard negatives, that is, documents that are highly confusable with the true positive. However, determining the optimal number of analyzed negatives ($k$) presents a critical trade-off. Inspecting too few candidates provides an insufficient signal to capture the model's precise confusion boundary. Conversely, expanding the pool too broadly risks diluting the diagnostic process with easily distinguishable, out-of-domain noise that distorts the semantic focus.

\begin{figure*}[!t] %
  \centering
  \includegraphics[width=\linewidth]{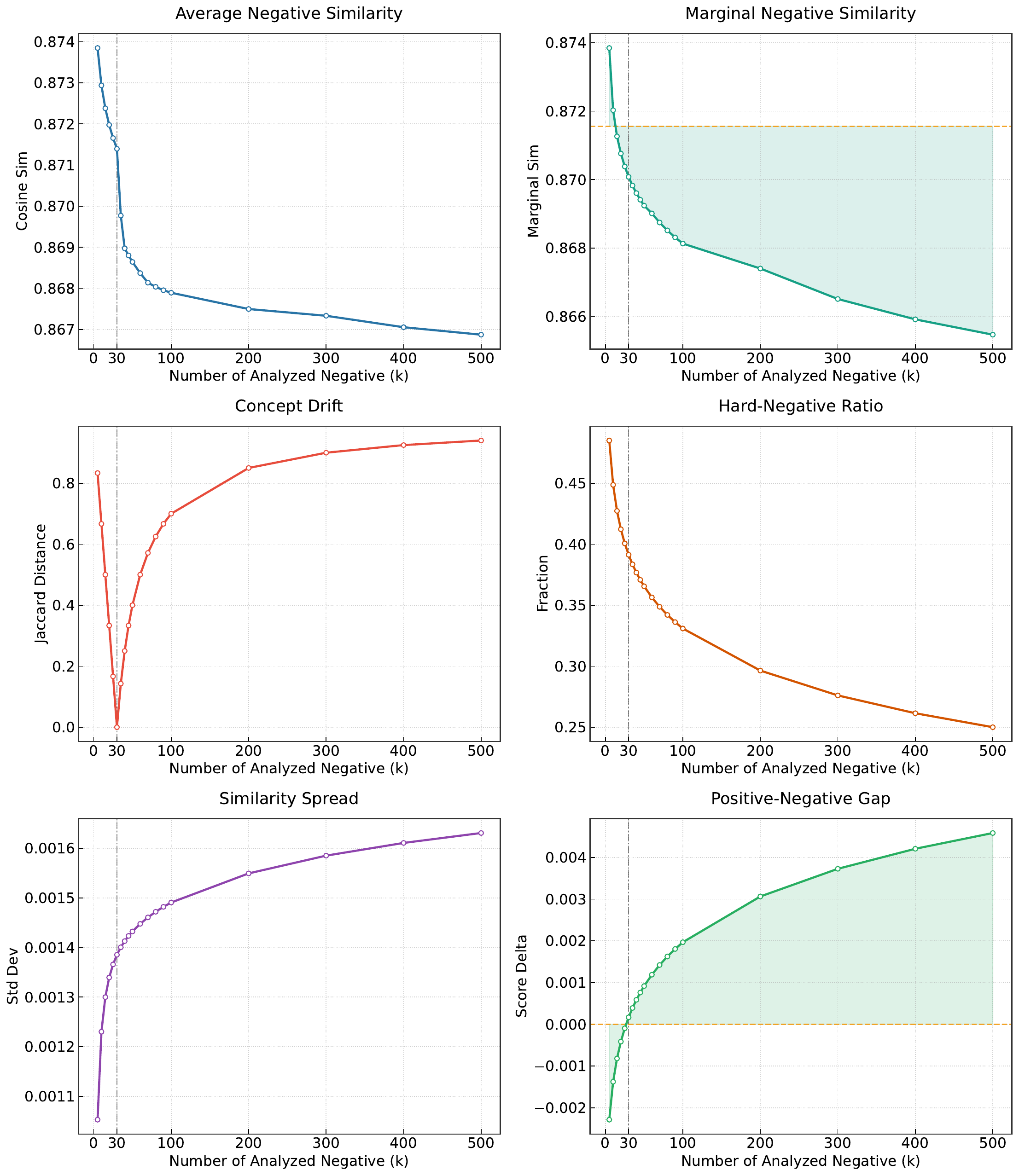}
  \caption{
    Fine-grained $k$ ablation ($k \le 100$) for the 0.5B Iter-0 retriever. The vertical dash-dot line marks the selected setting $k{=}30$.
  }
  \label{fig:appendix_k_fine}
\end{figure*}

To systematically justify our selection of $k=30$, we analyze the empirical results across six distinct metrics (visualized in Figure~\ref{fig:appendix_k_fine}). Rather than relying on arbitrary thresholds, this analysis demonstrates how $k=30$ provides an effective balance between maximizing diagnostic yield and mitigating semantic drift. For baseline comparisons, we define $\mathcal{N}_{30}(q)$ as the reference negative set.

\paragraph{Average Negative Similarity.}
To quantify how effectively the extracted distractors capture genuine confusion, we measure the average negative similarity. This metric reflects the overall difficulty of the negative pool, a higher value indicates that the retrieved documents remain highly competitive and structurally close to the query.

\begin{equation} 
\small
A_{\mathrm{avg}}(k) = \frac{1}{|\mathcal{Q}_{\mathrm{conf}}|}\sum_{q \in \mathcal{Q}_{\mathrm{conf}}} \frac{1}{k}\sum_{d \in \mathcal{N}_k(q)} s_\theta(q,d)
\end{equation}

As illustrated in Figure~\ref{fig:appendix_k_fine}(a), the pool maintains a high level of hardness up to $k=30$. Beyond this threshold, the similarity sharply declines, demonstrating that larger pools dilute the diagnostic quality with easily distinguishable documents. This validates $k=30$ as the optimal boundary for preserving concentrated hardness.

\paragraph{Marginal Negative Similarity.}
While the overall average similarity demonstrates general pool hardness, it can mask the diminishing quality of documents added at lower ranks. To isolate the exact diagnostic value of incrementally expanding the negative pool, we measure the marginal negative similarity. This metric specifically tracks the average similarity of the newly added documents between consecutive bounds $k_{\mathrm{prev}} < k$:

\begin{equation}
\small
\mu_k = \frac{1}{|\mathcal{Q}_{\mathrm{conf}}|}\sum_{q}\frac{1}{k-k_{\mathrm{prev}}} \sum_{j=k_{\mathrm{prev}}+1}^{k} s_\theta(q,\hat d_j)
\end{equation}

By comparing $\mu_k$ to the mean positive similarity ($\bar{s}^+ = \frac{1}{|\mathcal{Q}_{\mathrm{conf}}|}\sum_q s_\theta(q,d_q^+)$), we intuitively determine whether the freshly incorporated negatives are actually harder than the true positive. As illustrated in Figure~\ref{fig:appendix_k_fine}(b), once $k$ exceeds 30, $\mu_k$ drops significantly below the positive baseline. This confirms that documents ranked beyond 30 are, on average, easier for the model to distinguish than the true positive itself. Because they offer no meaningful diagnostic value, restricting the expansion to $k=30$ is strictly justified.

\paragraph{Concept Drift.}
To determine whether expanding the negative pool inadvertently introduces semantic noise, we quantify concept drift using the Jaccard distance relative to the $k=30$ reference set. This metric intuitively evaluates neighborhood stability, a value approaching 0 signifies strong alignment with the target semantic neighborhood, whereas higher values indicate substantial deviation.

\begin{equation}
\small
B_{\mathrm{drift}}(k) = \frac{1}{|\mathcal{Q}_{\mathrm{conf}}|}\sum_{q} \left(1 - \frac{|\mathcal{N}_k(q)\cap \mathcal{N}_{30}(q)|}{|\mathcal{N}_k(q)\cup \mathcal{N}_{30}(q)|}\right)
\end{equation}

As demonstrated in Figure~\ref{fig:appendix_k_fine}(c), the concept drift remains remarkably constrained up to $k=30$ but escalates rapidly thereafter. This sharp increase indicates that enlarging $k$ beyond 30 progressively pulls negatives from entirely different semantic neighborhoods, which compromises the precision of the diagnostic pool. Consequently, these results establish $k=30$ as the critical limit for maintaining semantic stability.

\paragraph{Hard-Negative Ratio.}
To assess the concentration of high-quality distractors within the pool, we calculate the hard-negative ratio. By defining a strict hardness threshold $\tau_h = A_{\mathrm{avg}}(5)$, this metric intuitively quantifies pool dilution; a lower $\rho_k$ implies that the retrieval space is saturated with easily distinguishable, non-informative documents.
\begin{equation}
\small
\rho_k = \frac{1}{|\mathcal{Q}_{\mathrm{conf}}|}\sum_{q} \frac{1}{k} \sum_{d \in \mathcal{N}_k(q)} \mathbb{I}[s_\theta(q,d) \ge \tau_h]
\end{equation}

As depicted in Figure~\ref{fig:appendix_k_fine}(d), maintaining $k=30$ preserves a dense fraction of effective distractors. Expanding the pool beyond this boundary results in severe dilution, establishing $k=30$ as the strict limit for maintaining the diagnostic quality of the negative set.

\paragraph{Similarity Spread.}
To observe the heterogeneity of the analyzed documents, we measure the similarity spread by computing the within-query standard deviation ($\sigma_k$) of the negative similarities. An increasing trend visually indicates a mixed pool of hard and easy documents rather than a dense, confusable cluster. Let $\bar{s}_k(q)$ be the average negative similarity for query $q$ within the top-$k$ pool, defined as $\frac{1}{k}\sum_{d'\in\mathcal{N}_k(q)}s_\theta(q,d')$. We measure the similarity spread $\sigma_k$ as the within-query standard deviation:
\begin{equation}
\small
\sigma_k = \frac{1}{|\mathcal{Q}_{\mathrm{conf}}|}\sum_{q} \sqrt{\frac{1}{k}\sum_{d\in\mathcal{N}_k(q)} \left(s_\theta(q,d) - \bar{s}_k(q)\right)^2}
\end{equation}

As shown in Figure~\ref{fig:appendix_k_fine}(e), the spread remains constrained up to $k=30$. Keeping the boundary here ensures the diagnosis mechanism focuses exclusively on a tightly packed cluster of errors.

\paragraph{Positive-Negative Gap.}
To directly quantify the degree of model confusion, we calculate the positive-negative gap, measuring the absolute difference between the true positive score and the average negative score. A value of $\gamma_k < 0$ highlights genuine confusion where negatives are scored higher than the positive.
\begin{equation}
\small
\gamma_k = \frac{1}{|\mathcal{Q}_{\mathrm{conf}}|}\sum_{q} \left(s_\theta(q,d_q^+) - \frac{1}{k}\sum_{d\in\mathcal{N}_k(q)} s_\theta(q,d)\right)
\end{equation}

As shown in Figure~\ref{fig:appendix_k_fine}(f), $\gamma_k$ becomes increasingly positive as $k$ grows past 30, meaning the average negative document becomes drastically easier than the positive document. This solidifies $k=30$ as the tipping point where true confusion is lost to general retrieval noise.

% =========================================================
\subsection{Effect of Fact-Verified Data Beyond Backbone Choice}
\label{app:cross_backbone}

To further verify that the gain stems from our data rather than a particular
backbone, we replace the backbone instead of the data. We apply the
\textsc{REPAIR} pipeline to the four backbones used by \textsc{BMRetriever}
(Pythia-410M, Pythia-1B, Gemma-2B, and BioMistral-7B), and compare each model
against the released \textsc{BMRetriever} model built on the same backbone.
Both models of a pair follow the setup of \S\ref{app:implementation} and are
evaluated on the five benchmarks of Table~\ref{tab:mainresult}. As shown in
Table~\ref{tab:cross_backbone_full}, \textsc{REPAIR} improves the average
nDCG@10 on all four backbones, by $+0.007$ to $+0.019$, and wins 18 of the 20
per-benchmark comparisons. The two exceptions both occur on Pythia-1B, where
SciFact ties and BIOSSES favors \textsc{BMRetriever}. This confirms that our
fact-verified data refinement drives the improvement across heterogeneous
backbone families, rather than benefiting from the specific capacity of
Qwen2.5.

\begin{table}[!t]
\centering
\arraybackslash
\fontsize{6.5}{7.5}\selectfont
\setlength{\tabcolsep}{3pt}
\renewcommand{\arraystretch}{1.6}
\resizebox{\columnwidth}{!}{%
\begin{tabular}{l|l|ccccc|c}
\hline
Backbone & Data & NFCorpus & SciFact & SciDocs & COVID & BIOSSES & AVG. \\ \hline
\multirow{2}{*}{Pythia-410M} & \textsc{BMRetriever} & 0.321 & 0.711 & 0.167 & 0.831 & 0.840 & 0.574 \\
 & {\cellcolor[HTML]{   EBF2F6}\textsc{REPAIR}} & {\cellcolor[HTML]{   EBF2F6}\textbf{0.339}} & {\cellcolor[HTML]{   EBF2F6}\textbf{0.728}} & {\cellcolor[HTML]{   EBF2F6}\textbf{0.183}} & {\cellcolor[HTML]{   EBF2F6}\textbf{0.847}} & {\cellcolor[HTML]{   EBF2F6}\textbf{0.848}} & {\cellcolor[HTML]{   EBF2F6}\textbf{0.589}} \\ \hline
\multirow{2}{*}{Pythia-1B} & \textsc{BMRetriever} & 0.344 & \textbf{0.760} & 0.180 & 0.840 & \textbf{0.858} & 0.596 \\
 & {\cellcolor[HTML]{   EBF2F6}\textsc{REPAIR}} & {\cellcolor[HTML]{   EBF2F6}\textbf{0.355}} & {\cellcolor[HTML]{   EBF2F6}\textbf{0.760}} & {\cellcolor[HTML]{   EBF2F6}\textbf{0.198}} & {\cellcolor[HTML]{   EBF2F6}\textbf{0.857}} & {\cellcolor[HTML]{   EBF2F6}0.852} & {\cellcolor[HTML]{   EBF2F6}\textbf{0.603}} \\ \hline
\multirow{2}{*}{Gemma-2B} & \textsc{BMRetriever} & 0.351 & 0.760 & 0.199 & 0.863 & 0.828 & 0.600 \\
 & {\cellcolor[HTML]{   EBF2F6}\textsc{REPAIR}} & {\cellcolor[HTML]{   EBF2F6}\textbf{0.372}} & {\cellcolor[HTML]{   EBF2F6}\textbf{0.778}} & {\cellcolor[HTML]{   EBF2F6}\textbf{0.218}} & {\cellcolor[HTML]{   EBF2F6}\textbf{0.876}} & {\cellcolor[HTML]{   EBF2F6}\textbf{0.840}} & {\cellcolor[HTML]{   EBF2F6}\textbf{0.617}} \\ \hline
\multirow{2}{*}{BioMistral-7B} & \textsc{BMRetriever} & 0.364 & 0.778 & 0.201 & 0.861 & 0.847 & 0.610 \\
 & {\cellcolor[HTML]{   EBF2F6}\textsc{REPAIR}} & {\cellcolor[HTML]{   EBF2F6}\textbf{0.388}} & {\cellcolor[HTML]{   EBF2F6}\textbf{0.798}} & {\cellcolor[HTML]{   EBF2F6}\textbf{0.225}} & {\cellcolor[HTML]{   EBF2F6}\textbf{0.876}} & {\cellcolor[HTML]{   EBF2F6}\textbf{0.857}} & {\cellcolor[HTML]{   EBF2F6}\textbf{0.629}} \\ \hline
\end{tabular}}
\caption{Experiments on the effect of fact-verified data across the four
backbones used by \textsc{BMRetriever}. Each pair trains the same backbone on
\textsc{BMRetriever}'s data and on ours. All scores are reported in nDCG@10.
The best-performing results within each backbone are highlighted in
\textbf{boldface}.}
\label{tab:cross_backbone_full}
\end{table}

% =========================================================
\subsection{Extended Iterations and Computational Cost}
\label{app:cost}

\paragraph{Saturation Beyond Two Iterations.}
Table~\ref{tab:abblationiter} extends the refinement loop to four iterations
at every model scale under the same $(p, k)$ setting. Moving from the second
to the third iteration raises the average nDCG@10 by $+0.005$ (500M),
$+0.003$ (1.5B) and $+0.003$ (7B), and a fourth iteration adds a further
$+0.001$ in all three cases, while each additional round consumes another
0.5M training pairs. The saturation point is thus the same across scales and
is reached without re-tuning $p$ or $k$, which is why we fix the number of
iterations to two throughout the paper.

\paragraph{Cost Structure.}
The cost profile of \textsc{REPAIR} differs structurally from that of
LLM-based augmentation. The Stage II API calls
(\textsc{Semantic Scholar}, \textsc{PubChem}, \textsc{MatProj}) are issued
offline in batch and are fully decoupled from the contrastive training loop,
so they consume no GPU time. Approaches that synthesize training data with a
generative model instead pay an inference cost at every augmentation step,
together with the downstream cost of filtering the hallucinations this
introduces~\cite{xu2024bmretriever, wang2024improving}. \textsc{REPAIR}
secures fact-verified evidence without incurring either.

\paragraph{Per-Stage Wall-Clock Cost.}
Table~\ref{tab:cost} itemizes the wall-clock cost of a single iteration for
each model scale, measured on two NVIDIA H200 GPUs under the training
configuration of \S\ref{app:implementation}. Even for the 7B model, one
iteration takes ${\sim}13$h, so the two iterations used throughout the paper
amount to ${\sim}52$ GPU-hours. For reference, RepLLaMA-7B reports four days
on $16{\times}$V100 (${\sim}1{,}500$ GPU-hours)~\cite{ma2024fine}, and
Promptriever-7B follows the same training
recipe~\cite{DBLP:conf/iclr/WellerDLPZH25}; fine-tuning E5-Mistral for
SFR-Embedding alone requires 120 GPU-hours (15h on
$8{\times}$A100)~\cite{meng2024sfr}, excluding its weakly-supervised
pre-training stage~\cite{wang2024improving}. The total compute of
\textsc{REPAIR} is thus one to two orders of magnitude below that of
comparable 7B-scale baselines.

\begin{table}[!t]
\centering
\fontsize{7}{8.5}\selectfont
\setlength{\tabcolsep}{3.5pt}
\renewcommand{\arraystretch}{1.2}
\resizebox{\columnwidth}{!}{%
\begin{tabular}{l|ccc|cc}
\hline
\multirow{2}{*}{Model} & Stage I & Stage II & Stage III & API Calls & Total \\
 & (Diag.) & (Expan.) & (Diff.) & (offline) & / Iter \\ \hline
\textsc{REPAIR}-500M & ${\sim}1.5$h & ${\sim}4$h & ${\sim}0.5$h & ${\sim}250$K & ${\sim}6$h \\
\textsc{REPAIR}-1.5B & ${\sim}2.5$h & ${\sim}4$h & ${\sim}1.0$h & ${\sim}250$K & ${\sim}7.5$h \\
\textsc{REPAIR}-7B   & ${\sim}6$h   & ${\sim}4$h & ${\sim}3$h   & ${\sim}250$K & ${\sim}13$h \\ \hline
\end{tabular}}
\caption{Per-iteration wall-clock cost of each \textsc{REPAIR} stage,
measured on $2{\times}$H200 GPUs. Stages I and III are GPU-bound, while Stage
II is bound by API latency and runs offline in batch without GPU cost. All
values are approximate.}
\label{tab:cost}
\end{table}
% =========================================================

% =========================================================
\subsection{Generalization to Additional Scientific Benchmarks}
\label{app:extra_bench}

The nine benchmarks of \S\ref{sec:ablation} already span five task families,
but they are drawn from materials science and biomedicine. To test whether
the long-tail resolution mechanism of \textsc{REPAIR} carries to task formats
and subareas it was never tuned for, we evaluate on three further scientific
benchmarks that appear nowhere in training: DORIS-MAE~\cite{wang2023doris},
whose queries are multi-aspect research summaries; CQA-physics~\cite{hoogeveen2015cqadupstack,
thakur2021beir}, whose queries are informal community posts from a physics
forum; and SciQ~\cite{welbl2017crowdsourcing}, which spans physics,
chemistry, biology, and earth science. Dataset statistics are given in
\S\ref{app:taskdataset}, and all scores are nDCG@10.

Table~\ref{tab:extra_bench} groups the results by parameter tier. Within
every tier \textsc{REPAIR} outperforms the corresponding \textsc{BMRetriever}
model, by $+0.040$ at 500M, $+0.005$ at 1.5B and $+0.022$ at 7B, and
\textsc{REPAIR}-7B attains the highest average overall ($0.6569$) despite
training on 4M pairs. The gains are largest on DORIS-MAE, where
\textsc{REPAIR} leads at all three tiers, indicating that resolving long-tail
confusion transfers to the multi-aspect query format the model never saw.
Two comparisons are closer. \textsc{REPAIR}-500M is on par with BGE-Large
($0.6143$ vs.\ $0.6149$), which is trained on 2.8B pairs, roughly $700\times$
our data; and \textsc{REPAIR}-1.5B trails \textsc{BMRetriever}-2B on SciQ
alone ($0.7944$ vs.\ $0.8013$) while remaining ahead on average. Overall,
performance holds up outside the domains and task formats the framework was
developed on.

\begin{table}[!t]
\centering
\fontsize{7.5}{9}\selectfont
\setlength{\tabcolsep}{4pt}
\renewcommand{\arraystretch}{1.15}
\resizebox{\columnwidth}{!}{%
\begin{tabular}{l|ccc|c}
\hline
Model & DORIS-MAE & CQA-physics & SciQ & AVG. \\ \hline
\textsc{BMRetriever}-410M & 0.5361 & 0.4113 & 0.7763 & 0.5746 \\
BGE-Large & 0.5947 & \textbf{0.4706} & 0.7793 & \textbf{0.6149} \\
\cellcolor[HTML]{   EBF2F6}\textsc{REPAIR}-500M (ours) & \cellcolor[HTML]{   EBF2F6}\textbf{0.6024} & \cellcolor[HTML]{   EBF2F6}0.4597 & \cellcolor[HTML]{   EBF2F6}\textbf{0.7809} & \cellcolor[HTML]{   EBF2F6}0.6143 \\ \hline
SGPT-2.7B & 0.5030 & 0.3240 & 0.6189 & 0.4820 \\
\textsc{BMRetriever}-2B & 0.5942 & 0.4742 & \textbf{0.8013} & 0.6232 \\
\cellcolor[HTML]{   EBF2F6}\textsc{REPAIR}-1.5B (ours) & \cellcolor[HTML]{   EBF2F6}\textbf{0.6140} & \cellcolor[HTML]{   EBF2F6}\textbf{0.4773} & \cellcolor[HTML]{   EBF2F6}0.7944 & \cellcolor[HTML]{   EBF2F6}\textbf{0.6286} \\ \hline
E5-Mistral & 0.5308 & 0.4950 & 0.7954 & 0.6071 \\
\textsc{BMRetriever}-7B & 0.6241 & 0.4622 & 0.8197 & 0.6353 \\
\cellcolor[HTML]{   EBF2F6}\textsc{REPAIR}-7B (ours) & \cellcolor[HTML]{   EBF2F6}\textbf{0.6289} & \cellcolor[HTML]{   EBF2F6}\textbf{0.5051} & \cellcolor[HTML]{   EBF2F6}\textbf{0.8366} & \cellcolor[HTML]{   EBF2F6}\textbf{0.6569} \\ \hline
\end{tabular}}
\caption{Experiments on three additional scientific benchmarks that are
held out from training, grouped by parameter scale. All scores are reported
in nDCG@10 and given to four decimals, since several comparisons differ only
in the fourth. The best-performing results within each scale are highlighted
in \textbf{boldface}.}
\label{tab:extra_bench}
\end{table}
% =========================================================

% =========================================================
\subsection{Citation Adjacency of Mined Hard Negatives}
\label{app:citation}

Table~\ref{tab:citation_rate} reports the citation check behind the claim in
\S\ref{sec:ablation}. For each source we sample 500 anchor documents, look up
every pair in \textsc{Semantic Scholar}, and count a pair as adjacent if
either document cites the other. Two reference points frame the result.
Randomly paired documents give $0.000\%$, the floor of the measurement.
SciDocs positive pairs, which are built from citation links and should
therefore give $100\%$, give only $5.80\%$: the lookup finds a citation for
just 29 of 500 pairs, because \textsc{Semantic Scholar} indexes few
references for older papers. This $5.80\%$ is thus the highest rate the check
can return, not the true rate. The hard negatives mined by \textsc{REPAIR}
sit below $0.00001\%$, far closer to the random floor than to this ceiling,
so the documents our diagnosis treats as negatives are almost never
overlooked positives.

\begin{table}[!t]
\centering
\fontsize{7.5}{9}\selectfont
\setlength{\tabcolsep}{5pt}
\renewcommand{\arraystretch}{1.15}
\resizebox{\columnwidth}{!}{%
\begin{tabular}{l|cc}
\hline
Pair source & Citation rate & Matched / total \\ \hline
Random pairs & $0.000\%$ & 0 / 10,623 \\
\textsc{REPAIR} hard negatives & $<0.00001\%$ & -- \\
SciDocs positive pairs & $5.80\%$ & 29 / 500 \\ \hline
\end{tabular}}
\caption{Citation rates of three sources of document pairs, measured with
the same \textsc{Semantic Scholar} lookup. SciDocs positive pairs are already
linked by citation, so their $5.80\%$ is the highest rate the lookup can
detect rather than a true rate.}
\label{tab:citation_rate}
\end{table}
% =========================================================

% =================================================================================================================

%=============================
%=============================
\begin{table*}[!t]
\centering
\small
\renewcommand{\arraystretch}{1.3}
\begin{tabular}{p{0.15\linewidth}|p{0.2\linewidth}|p{0.25\linewidth}|p{0.25\linewidth}}
\hline
\textbf{Error Category} & \textbf{Generated Query} & \textbf{Positive Document} & \textbf{Negative Document} \\ \hline

\textbf{Case 1:} \newline Entity Number Swap & 
which myeloma cell line carries prosurvival \textcolor{red}{bcl1} & 
... Myeloma cells usually express a range of the prosurvival \textcolor{red}{BCL2} proteins. ... & 
... MCL1, an anti-apoptotic \textcolor{red}{BCL2} family protein, is a key regulator ... \\ \hline

\textbf{Case 2:} \newline Fact Direction Reversal & 
is MAU associated with a \textcolor{red}{decreased} recurrence of cardiovascular events & 
... Microalbuminuria (MAU) is associated with an \textcolor{red}{enhanced} risk of cardiovascular events. ... & 
... Microalbuminuria (MA) is a known marker for endothelial dysfunction and future cardiovascular \textcolor{red}{events}. ... \\ \hline

\textbf{Case 3:} \newline Disease Entity Confusion & 
what is transpyloric shuttle in \textcolor{red}{diabetes} & 
... The TransPyloric Shuttle (TPS) is a nonsurgical device ... to treat \textcolor{red}{obesity}. ... & 
... \textcolor{red}{Diabetes} distress (DD), or psychological fatigue associated with \textcolor{red}{diabetes} management ... \\ \hline

\textbf{Case 4:} \newline Chemical Substitution & 
does \textcolor{red}{fructose} cause vesicles to accumulate? & 
... treatment of mammalian cells with \textcolor{red}{sucrose} leads to vacuole accumulation ... & 
... Increased \textcolor{red}{fructose} concentrations are the biochemical hallmark of \textcolor{red}{fructosemia} ... \\ \hline

\textbf{Case 5:} \newline Kinase Variant Swap & 
does \textcolor{red}{CDK6} suppress TSC2 phosphorylation? & 
... cyclin D1/\textcolor{red}{CDK4} mediate resistance ... Inhibition of \textcolor{red}{CDK4/6} ... reduces TSC2 phosphorylation ... & 
... Targeting cyclin-dependent kinases \textcolor{red}{4/6} (CDK4/6) represents a therapeutic option ... \\ \hline

\end{tabular}
\caption{Examples of LLM-generated dataset errors. Red text indicates hallucinated entities, misattributed scientific facts, or context stripped from negative documents.}
\label{tab:hallucination_analysis_detailed}
\end{table*}
%=============================
%---------------

%=============================
% Appendix: Analysis of LLM-Generated Dataset Errors
%=============================

\section{Analysis of LLM-Generated Dataset Errors}
\label{sec:appendix_error_analysis}

Table~\ref{tab:hallucination_analysis_detailed} presents representative failure cases identified from a qualitative inspection of a subset of the synthetic dataset\footnote{\url{https://huggingface.co/datasets/BMRetriever/biomed_retrieval_dataset}} generated by existing LLM-based augmentation methods \cite{xu2024bmretriever}. Although our analysis is confined to a limited sample, the severity and fundamental nature of the uncovered errors suggest a risk that such structural hallucinations may be present throughout the corpus. A more comprehensive investigation is warranted to determine the full spectrum of these critical flaws. While naive prompt-based generation has shown empirical success in general-domain retrieval, our findings reveal that current LLMs fundamentally struggle with the \textbf{long-tailed concept distribution (P1)} and \textbf{high fact-sensitivity (P2)} of scientific texts. This limitation inevitably leads to the generation of harmful, hallucinatory data that degrades retriever performance. Based on our manual review, we categorize the observed vulnerabilities into four primary failure modes, explicitly highlighting why our proposed methodology is strictly necessary to overcome these bottlenecks.

\subsection*{Failure Mode 1: Entity Number and Sub-variant Swap (P1 \& P2)}
LLMs frequently treat structurally similar but biologically distinct entities as interchangeable tokens, especially within long-tailed biomedical concepts. 
\begin{itemize}
    \item \textbf{Analysis of Case 1 \& 5:} In Case 1, the LLM confuses \textcolor{red}{BCL1} with \textcolor{good}{BCL2} under the exact same context of "prosurvival myeloma proteins." Similarly, in Case 5, \textcolor{red}{CDK6} is swapped with \textcolor{good}{CDK4}. To a general-domain LLM, a single-digit difference represents a negligible semantic shift. However, in the biomedical domain, this minor perturbation completely invalidates the scientific fact.
    \item \textbf{Why our method is required:} Naive generative models cannot self-correct these single-token factual violations. Our methodology specifically addresses this by enforcing strict entity-grounding constraints, ensuring that long-tailed numerical variants are perfectly aligned between the query and the positive document.
\end{itemize}

\subsection*{Failure Mode 2: Fact Direction Reversal (P2)}
Medical literature is highly sensitive to the directionality of outcomes (e.g., increase vs. decrease, inhibit vs. promote). LLMs often hallucinate these directional markers because opposite terms frequently co-occur in similar training contexts.
\begin{itemize}
    \item \textbf{Analysis of Case 2:} The generated query asks about a \textcolor{red}{decreased} cardiovascular risk associated with Microalbuminuria (MAU), whereas the positive document explicitly states an \textcolor{good}{enhanced} risk. The LLM successfully grasped the topic (MAU and cardiovascular risk) but completely inverted the medical conclusion.
    \item \textbf{Why our method is required:} This demonstrates that semantic similarity alone is insufficient for scientific retrieval. Our approach directly addresses High Fact-Sensitivity (P2) by verifying the causal and directional consistency of the generated triplets, preventing the model from learning biologically fatal contradictions.
\end{itemize}

\subsection*{Failure Mode 3: Disease Entity Confusion via Context Stripping}
LLMs often suffer from attention leakage when processing multiple documents, mistakenly integrating concepts from negative documents into the query intended for the positive document.
\begin{itemize}
    \item \textbf{Analysis of Case 3:} The positive document describes a device to treat \textcolor{good}{obesity}. However, the LLM inserts \textcolor{red}{diabetes} into the query. This hallucination occurs because the surrounding negative documents (or the LLM's internal prior) strongly associate obesity treatments with diabetes, causing a cross-contamination of concepts.
    \item \textbf{Why our method is required:} This proves that providing LLMs with negative documents as prompt context often degrades query quality rather than improving it. Our pipeline introduces a robust isolation mechanism that prevents negative context bleeding, maintaining the exact conceptual boundaries of the target document.
\end{itemize}

\subsection*{Failure Mode 4: Chemical Substitution}
Similar to numerical swaps, LLMs fail to distinguish between fundamental chemical compounds that share functional or structural categories.
\begin{itemize}
    \item \textbf{Analysis of Case 4:} The LLM replaces \textcolor{good}{sucrose} with \textcolor{red}{fructose}. While both are sugars, the specific vacuole accumulation process described in the document is exclusive to sucrose in this experimental context. 
    \item \textbf{Why our method is required:} Our proposed filtering and generation strategy explicitly penalizes out-of-context chemical substitutions. By leveraging domain-specific hard-negative mining, we force the retriever to learn the precise distinctions between such granular entities, a capability entirely absent in datasets generated by baseline LLM approaches.
\end{itemize}

\paragraph{Conclusion on Novelty}
The examples delineated in Table~\ref{tab:hallucination_analysis_detailed} are not mere edge cases; they are systemic failures stemming from the inherent architectural limitations of unconstrained LLMs. Generating training data with these undetected hallucinations forces retrieval models to learn scientifically false representations. The novelty of our proposed methodology lies in its structural capability to categorically eliminate these failure modes, specifically addressing long-tailed entity swaps and fact-direction reversals, thereby producing a high-fidelity, factually rigorous dataset that significantly elevates biomedical retrieval performance.
%

%=================================

\begin{table*}[!t]
\centering

\renewcommand{\arraystretch}{1.3}
\small
\begin{tabular}{@{}p{2cm} p{3.3cm} p{1.5cm} p{6.5cm} c@{}}
\toprule
\textbf{Dataset} & \textbf{User Query} & \textbf{Model} & \textbf{Top-1 Retrieved Snippet (Truncated)} & \textbf{Match} \\
\midrule

% Case 1: SciFact
\multirow{3}{2.2cm}{\textbf{SciFact}\newline (Biology/Fact)} 
& \multirow{3}{3.8cm}{Less than 10\% of the gabonese children with SFM had a plasma lactate of more than 5mmol/L.} 
& \textbf{REPAIR} & \textbf{[Correct]} ...measured body compartment volumes in \textbf{Gabonese children} with malaria... & \textbf{O} \\ \cmidrule(l){3-5} 
& & BMR & \textit{[Irrelevant]} Compound heterozygous ZMPSTE24 mutations reduce prelamin A processing... & X \\ \cmidrule(l){3-5} 
& & E5M & \textit{[Lexical Trap]} \textbf{Lactic} acidosis in patients with \textbf{diabetes} treated with metformin... & X \\ 
\midrule

% Case 2: NFCorpus
\multirow{3}{2.2cm}{\textbf{NFCorpus}\newline (Nutrition)} 
& \multirow{3}{3.8cm}{red tea} 
& \textbf{REPAIR} & \textbf{[Correct]} ...elucidate health benefit of \textbf{herbal teas}... \textbf{green tea}, \textbf{black tea}... & \textbf{O} \\ \cmidrule(l){3-5} 
& & BMR & \textit{[Lexical Trap]} Color \textbf{red} reduces snack food soft drink intake... & X \\ \cmidrule(l){3-5} 
& & E5M & \textit{[Partial]} ...antimutagenic activity \textbf{white tea} comparison green tea... & X \\ 
\midrule

% Case 3: ChemLit
\multirow{3}{2.2cm}{\textbf{ChemLit}\newline (Chemistry Proc.)} 
& \multirow{3}{3.8cm}{What is the step before heating the solution in the process?} 
& \textbf{REPAIR} & \textbf{[Correct]} ...The \textbf{Teflon screw top was closed} on the J-young NMR tube, and the \textbf{solution was heated}... & \textbf{O} \\ \cmidrule(l){3-5} 
& & BMR & \textit{[Irrelevant]} Treatment of [TpMo(CO)3] with 1 equiv of gray Se in THF-d8... failed to produce... & X \\ \cmidrule(l){3-5} 
& & E5M & \textit{[Partial]} ...solution was heated at 50 °C on a hot plate... turned from yellow to dark red... & $\triangle$ \\

\bottomrule
\end{tabular}

\caption{Comparative Case Study of Retrieval Performance Across Diverse Domains}
\label{tab:case_study}
\end{table*}

\section{Extended Case Study Results}
\label{sec:appendix}

\subsection{Qualitative Analysis of Retrieval Capabilities}
\label{sec:appendix_case_study}

To explicitly demonstrate the superiority of the REPAIR framework over existing strong dense retrieval baselines, \textsc{BMRetriever}~\cite{xu2024bmretriever} and E5-Mistral~\cite{wang2024improving}, we present an in-depth qualitative comparison. We specifically targeted three highly specialized domains that challenge distinct retrieval capabilities: biomedical fact-verification (SciFact), nutritional literature (NFCorpus), and chemical procedural reasoning (ChemLit). As illustrated in Table \ref{tab:case_study}, conventional models frequently fall into the trap of superficial lexical overlap or fail to capture complex relational logic. In contrast, REPAIR successfully isolates deep semantic structures, factual nuances, and procedural causality. This robustness directly stems from our self-evolving methodology, which trains the model to comprehend holistic context rather than relying on token-level matching.

\paragraph{Case 1: Resolving Complex Factual Constraints (SciFact).}
In the SciFact example, the user query demands the precise intersection of demographic data ("Gabonese children") and clinical measurements ("plasma lactate"). While E5-Mistral is completely derailed by the keyword "lactic" and retrieves an irrelevant document about lactic acidosis in diabetes (a classic lexical trap), REPAIR accurately localizes the specific demographic and clinical context. This highlights REPAIR's novelty in maintaining multi-hop factual integrity without being distracted by high-frequency medical jargon.

\paragraph{Case 2: Ontological Understanding over Lexical Matching (NFCorpus).}
The "red tea" query exposes the limitations of traditional semantic models in handling ambiguous, real-world terms. \textsc{BMRetriever} erroneously focuses on the exact color "red" in an entirely unrelated context (snack food packaging). Conversely, REPAIR exhibits a sophisticated understanding of ontological categories, successfully retrieving documents conceptually mapped to "herbal teas," "green tea," and "black tea." This demonstrates REPAIR's capability to map queries to broader semantic clusters, proving its effectiveness in domains where exact keyword overlaps are sparse.

\paragraph{Case 3: Procedural and Temporal Reasoning (ChemLit).}
Perhaps the most striking evidence of REPAIR's novelty lies in the ChemLit domain, which strictly requires sequential reasoning. The query explicitly asks for the step \textit{before} a specific action ("heating the solution"). While E5-Mistral retrieves a snippet that simply describes the heating process (a partial match that entirely misses the temporal prerequisite), REPAIR accurately identifies the chronological predecessor ("The Teflon screw top was closed"). This proves that REPAIR goes beyond static semantic matching to comprehend dynamic, procedural causality, a significant and novel advancement over current baseline models.

\section{Robustness to Concept Extraction Noise}
\label{app:concept_overlap}

A fundamental strength of the proposed REPAIR framework is its capacity for continuous epistemic renewal. Rather than stagnating in a self-reinforcing feedback loop of existing model biases, the iterative refinement process dynamically resolves prior confusions while continuously uncovering novel epistemic boundaries. This structural advantage is guaranteed by the Expansion stage, which anchors newly diagnosed concepts in externally verified knowledge bases (e.g., Semantic Scholar, \textsc{PubChem}, \textsc{MatProj}) rather than relying solely on internal model-generated distributions.

To empirically validate this dynamic self-correction and demonstrate that the model does not merely reinforce its own bias, we analyze the evolution of the diagnosed confused concept set ($\mathcal{C}_\text{conf}$) and the confusion query set ($\mathcal{Q}_\text{conf}$) across consecutive iterations. We track the transition from the seed model on the initial corpus $\mathcal{T}_0$ (Iteration 1) to the refined model on the augmented corpus $\mathcal{T}_1$ (Iteration 2) using the REPAIR-500M setup. Both iterations employ a selection ratio of $p=40\%$ and $k=30$ negatives. We utilize three key metrics to capture the nature of this representational shift:

\paragraph{Concept-Level Set Overlap (Jaccard Similarity).}
We first investigate whether the model is simply trapped in a cycle of repeating its past mistakes. To quantify this, we calculate the Jaccard similarity between the confused concept set from Iteration 1 ($\mathcal{C}_\text{conf}^{(1)}$) and Iteration 2 ($\mathcal{C}_\text{conf}^{(2)}$).

Intuitively, if the refinement process were merely reinforcing existing biases, we would observe a high overlap; this would indicate that the model continually struggles with the exact same concepts (epistemic stagnation). Conversely, a low overlap demonstrates that the model successfully resolves past confusions and progresses to discover new, uncharted boundaries.

As shown in Table~\ref{tab:app_set_overlap}, the Jaccard similarity is remarkably low at $0.146$. This low overall overlap is driven by two highly positive outcomes: first, nearly half ($48.6\%$) of the concepts that confused the Iteration-1 model are completely resolved after just one refinement step. Second, the vast majority ($83.1\%$) of the concepts diagnosed in Iteration 2 are entirely novel. Together, these statistics provide clear evidence that the model is actively expanding its knowledge rather than stagnating in a feedback loop.

\begin{table}[!t]
\centering
\small

\begin{tabular}{lrr}
\toprule
\textbf{Statistic} & \textbf{Count} & \textbf{Percentage} \\
\midrule
$|\mathcal{C}_\text{conf}^{(1)}|$ & 531,305 & -- \\
$|\mathcal{C}_\text{conf}^{(2)}|$ & 1,613,447 & -- \\
\midrule
Persistent ($\mathcal{C}^{(1)} \cap \mathcal{C}^{(2)}$) & 273,199 & 14.6\% of union \\
Resolved ($\mathcal{C}^{(1)} \setminus \mathcal{C}^{(2)}$) & 258,106 & 48.6\% of Iter-1 \\
Novel ($\mathcal{C}^{(2)} \setminus \mathcal{C}^{(1)}$) & 1,340,248 & 83.1\% of Iter-2 \\
\midrule
\textbf{Jaccard similarity} & \multicolumn{2}{c}{$\mathbf{0.146}$} \\
\bottomrule
\end{tabular}

\caption{Concept set overlap between $\mathcal{C}_\text{conf}^{(1)}$ and $\mathcal{C}_\text{conf}^{(2)}$, REPAIR-500M.}
\label{tab:app_set_overlap}
\end{table}

\paragraph{Top-$K$ Severity Persistence and Rank Correlation.}
Beyond general set overlap, it is critical to determine whether the most severe confusions persist. If a bias feedback loop were active, the highest-ranked confusion targets (measured by CCS score) would remain anchored at the top of the distribution. Table~\ref{tab:app_topk_overlap} demonstrates that the Jaccard similarity for the top-100 highest-CCS concepts is strictly zero. Extending this observation to the top-1,000 yields a near-zero similarity of $0.003$. Furthermore, among the fractional subset of concepts that do persist across both iterations, their severity ordering is fundamentally disrupted; the Spearman rank correlation ($\rho$) of their CCS scores is merely $0.111$ (Table~\ref{tab:app_ccs_shift}). This confirms that the refinement process decisively dismantles the most severe representational bottlenecks.

\begin{table}[!t]
\centering
\small

\begin{tabular}{rrrr}
\toprule
\textbf{Top-$K$} & \textbf{Persistent} & \textbf{Novel (Iter-2)} & \textbf{Jaccard} \\
\midrule
100     & 0       & 100     & 0.0000 \\
500     & 1       & 499     & 0.0010 \\
1,000   & 6       & 994     & 0.0030 \\
5,000   & 39      & 4,961   & 0.0039 \\
10,000  & 111     & 9,889   & 0.0056 \\
50,000  & 1,640   & 48,360  & 0.0167 \\
100,000 & 5,512   & 94,488  & 0.0283 \\
\bottomrule
\end{tabular}
\caption{Top-$K$ concept Jaccard by CCS rank, REPAIR-500M.}
\label{tab:app_topk_overlap}
\end{table}

\begin{table}[!t]
\centering
\small

\begin{tabular}{lrrl}
\toprule
\textbf{Metric} & \textbf{Iter-1} & \textbf{Iter-2} & \textbf{$\Delta$} \\
\midrule
CCS Median     & 2.09  & 3.89  & $+1.80$ \\
CCS Mean       & 6.88  & 5.59  & $-1.28$ \\
Spearman $\rho$ (CCS rank) & \multicolumn{3}{c}{$\mathbf{0.111}$} \\
\bottomrule
\end{tabular}
\caption{CCS statistics for persistent concepts ($\mathcal{C}^{(1)} \cap \mathcal{C}^{(2)}$), REPAIR-500M.}
\label{tab:app_ccs_shift}
\end{table}
\paragraph{Query-Level Margin Shift.}
Finally, we track the evolutionary trajectory at the query level. The query Jaccard similarity stands at $0.0001$ (Table~\ref{tab:app_query_overlap}), indicating that the augmented corpus $\mathcal{T}_1$ successfully provides the necessary supervision to resolve nearly all queries that confused the Iteration-1 model. Crucially, we isolate the behavior of the 173 persistent queries that remain in the confused set during Iteration 2. For this specific subset, we observe a positive margin shift from $-2.7 \times 10^{-3}$ to $+3.5 \times 10^{-3}$. This metric directly illustrates that even when a query necessitates multiple refinement rounds, the model's representational margins are actively expanding and separating, firmly countering any hypothesis of biased stagnation.

\begin{table}[!t]
\centering
\small
\resizebox{\columnwidth}{!}{%
\begin{tabular}{lrrr}
\toprule
\multirow{2}{*}{\textbf{Query Group}} & \multirow{2}{*}{\textbf{Count}} & \multicolumn{2}{c}{\textbf{Avg. Margin ($\times 10^{-3}$)}} \\
\cmidrule(lr){3-4}
& & \textbf{Iter-1} & \textbf{Iter-2} \\
\midrule
Persistent ($\mathcal{Q}^{(1)} \cap \mathcal{Q}^{(2)}$) & 173 & $-2.7$ & $+3.5$ \\
Resolved ($\mathcal{Q}^{(1)} \setminus \mathcal{Q}^{(2)}$) & 504,937 & $-4.4$ & -- \\
Novel ($\mathcal{Q}^{(2)} \setminus \mathcal{Q}^{(1)}$) & 745,843 & -- & $+7.2$ \\
\midrule
\textbf{Query Jaccard} & \multicolumn{3}{c}{\textbf{0.0001}} \\
\bottomrule
\end{tabular}%
}
\caption{$\mathcal{Q}_\text{conf}$ overlap and average margin shift across iterations for REPAIR-500M ($p=40\%$).}
\label{tab:app_query_overlap}
\end{table}

\end{document}